\documentclass[11pt]{article}
\pdfoutput=1
\usepackage[preprint]{acl}

\usepackage{times}
\usepackage{latexsym}

\usepackage{amsmath,amsfonts,bm}

\def\eqref#1{equation~\ref{#1}}

\def\1{\bm{1}}

\def\ervv{{\textnormal{v}}}

\DeclareMathAlphabet{\mathsfit}{\encodingdefault}{\sfdefault}{m}{sl}
\SetMathAlphabet{\mathsfit}{bold}{\encodingdefault}{\sfdefault}{bx}{n}

\def\gG{{\mathcal{G}}}

\newcommand{\parents}{Pa} 

\usepackage{hyperref}
\usepackage{url}

\usepackage{multirow}
\usepackage{booktabs}
\usepackage{graphicx}
\usepackage{enumitem}
\usepackage{amssymb}
\usepackage{color}

\usepackage{tcolorbox}
\definecolor{xm-purple}{RGB}{210, 210, 210}
\definecolor{xm-grey}{RGB}{242,242,242}
\newtcolorbox[list inside=prompt,auto counter]{prompt}[1][]{
    colbacktitle=xm-purple!90,
    colback =xm-grey!30,
    coltitle=black,
    fontupper=\footnotesize,
    boxsep=5pt,
    left=0pt,
    right=0pt,
    top=0pt,
    bottom=0pt,
    boxrule=0.5pt,
    #1,
}

\setlist{leftmargin=12pt}
\setlist[itemize]{itemsep=5pt, parsep=0pt, topsep=0pt, nosep}

\usepackage[T1]{fontenc}

\usepackage[utf8]{inputenc}

\usepackage{microtype}

\usepackage{inconsolata}

\usepackage{graphicx}

\title{ParaCook: On Time-Efficient Planning for Multi-Agent Systems}

\author{Shiqi Zhang$^{1}$\thanks{Equal Contribution. }, Xinbei Ma$^{1}$$^*$, Yunqing Xu$^{1}$, Zouying Cao$^{1}$, Pengrui Lu$^{1}$, \\
\textbf{Haobo Yuan$^{2}$, Tiancheng Shen$^{2}$, Zhuosheng Zhang$^{1}$, Hai Zhao$^{1}$, Ming-Hsuan Yang$^{2}$}
 \\
$^{1}$Shanghai Jiao Tong University $^{2}$University of California, Merced \\
\texttt{\{zsq259, sjtumaxb, xuyunqing, zouyingcao, lupengrui, zhangzs\}@sjtu.edu.cn}, \\
\texttt{zhaohai@cs.sjtu.edu.cn}, \texttt{\{haoboyuan, tianchengshen, mhyang\}@ucmerced.edu}, \\
}

\begin{document}
\maketitle

\newcommand{\paraplan}[0]{parallel planning}
\newcommand{\sas}[0]{single-agent system}
\newcommand{\mas}[0]{multi-agent system}
\newcommand{\as}[0]{agent team}

\newcommand{\gemini}[0]{\textit{Gemini-2.5-Pro}}
\newcommand{\gptfive}[0]{\textit{GPT-5}}
\newcommand{\ds}[0]{\textit{DeepSeek-V3.2-Exp}}
\newcommand{\claudeopusfourone}[0]{\textit{Claude-Opus-4.1}}
\newcommand{\qwen}[0]{\textit{Qwen3-Max-Preview}}

\newcommand{\cmark}{$\checkmark$}
\newcommand{\xmark}{$\times$}

\newcommand{\cotdelta}[2]{%
    \ifdim #1pt > #2pt 
        $#1_{\downarrow}$%
    \else\ifdim #1pt < #2pt 
        $#1_{\uparrow}$%
    \else 
        $#1$%
    \fi\fi
}

\newcommand{\cotdeltapoct}[2]{%
    \ifdim #1pt < #2pt 
        $#1_{\downarrow}$%
    \else\ifdim #1pt > #2pt 
        $#1_{\uparrow}$%
    \else 
        $#1$%
    \fi\fi
}
\begin{abstract}
Large Language Models (LLMs) exhibit strong reasoning abilities for planning long-horizon, real-world tasks, yet existing agent benchmarks focus on task completion while neglecting time efficiency in parallel and asynchronous operations. To address this, we present ParaCook, a benchmark for time-efficient collaborative planning. Inspired by the Overcooked game, ParaCook provides an environment for various challenging interaction planning of multi-agent systems that are instantiated as cooking tasks, with a simplified action space to isolate the core challenge of strategic \paraplan{}. Through a comprehensive evaluation of state-of-the-art LLMs, we find that current approaches achieve suboptimal plans, which struggle with parallel actions or coordination. Our analysis also reveals LLMs' potential on abstract tasks where they can focus on high-level parallel optimization. ParaCook provides a scalable evaluation framework with adjustable complexity, establishing a foundation for developing and assessing time efficiency-aware multi-agent planning.
The code and data are available at \url{https://github.com/zsq259/ParaCook}.
\end{abstract}

\section{Introduction}

Large Language Models (LLMs) have empowered agents with remarkable \textbf{planning} capabilities in complex and interactive tasks \cite{yang2025plan, shinn2023reflexion}.
Planning integrates task knowledge and breaks down long-horizon goals into subtasks, enabling overall coherence rather than just local optimality.
However, in long-horizon tasks with explicit temporal dynamics and concurrency, task correctness alone is no longer sufficient to characterize planning quality.
Even when all required subtasks are completed successfully, different plans may exhibit vastly different execution times due to idle periods, blocking, and resource contention.
As a result, effective planning in such settings fundamentally requires reasoning about \emph{time-efficient scheduling}, rather than correctness alone.

This motivates the research question that
\textbf{long-horizon tasks should not only be distributed correctly, but also scheduled efficiently}.
Efficient scheduling is non-trivial because complex tasks inherently involve both parallel and sequential steps.
For a single agent, schedules can exploit intra-agent parallelism between sub-tasks that do not block each other to reduce idle time. For example, during water boiling, the agent can switch to another action, like chopping vegetables.
For \mas{}, different agents can further leverage inter-agent parallelism by distributing independent subtasks, such as cooking different dishes simultaneously.
Crucially, these two forms of parallelism jointly determine planning quality under parallel execution, where multiple success-equivalent plans may differ significantly in overall completion time.


From this perspective, existing agent benchmarks are primarily designed to evaluate objectives orthogonal to time-efficient parallel scheduling.
Many focus on task success, rule-following, or coordination feasibility, without explicitly modeling execution time or shared-resource contention.
Other benchmarks consider concurrency under highly simplified assumptions.
AsyncHow \citep{lingraph} reduces planning to ordering pre-decomposed subtasks with given durations and dependency graphs, rather than execution-grounded scheduling.
Robotouille \citep{gonzalez-pumariega2025robotouille} studies asynchronous execution but is limited to single-agent planning.
CookBench \citep{cai2025cookbench} provides long-horizon embodied tasks, but its comprehensive visual and interaction complexity makes it difficult to isolate and evaluate scheduling efficiency.
As a result, current benchmarks do not fully expose opportunities for parallelism, leaving a notable gap in evaluating whether LLMs can truly exploit concurrency to minimize overall task completion time.


To address this gap, we propose \textbf{ParaCook}, a benchmark designed to evaluate \textit{time-efficient parallel scheduling} in embodied multi-agent planning.
We focus on cooking, an everyday scenario that naturally involves both sequential and parallel tasks.
Inspired by the Overcooked game, ParaCook simplifies action spaces and task rules, allowing models to concentrate on effective task scheduling and optimal time utilization. Our benchmark supports multiple agents cooperating, emphasizing both inter-agent parallelism and intra-agent time utilization. To further isolate planning from low-level actions, we introduce corresponding abstract tasks that focus solely on high-level task allocation and optimization of execution time,
enabling direct evaluation of LLMs’ planning capabilities and preserve the challenges of efficient scheduling.

Through comprehensive experiments, we find that top models like GPT-5 achieve only a 65\% average success rate with significant performance drops on complex tasks, while humans maintain perfect success. Models also exhibit substantially longer completion times and higher movement costs than human baselines. On abstract planning tasks, top LLMs achieve near-optimal performance (within 1-7\% of optimum), demonstrating strong high-level reasoning capabilities.

Our contributions can be summarized as follows: (i) ParaCook, the first benchmark for time-efficient multi-agent planning with systematic parallelism evaluation; (ii) A scalable framework with adjustable complexity control; (iii) Comprehensive analysis revealing critical gaps between current LLMs and human performance.

\section{Related Work}
In this section, we introduce studies on agents' planning, their extension to multi-agent cooperation, and benchmarks for evaluating planning.

\subsection{Planning for Agents}

LLMs have emerged as powerful planners due to their strong reasoning capabilities.

\textbf{Single-agent system planning.}
Planning enables an agent to decompose complex tasks and refine actions through iterative reasoning \cite{yao2023react, yang2025plan}.  
Mainstream studies focus on task decomposition \cite{shen2023hugginggpt, wang2023plan, gao2023pal,chen2022program},
or self-improvement with reflection on historical error in memory \citep{shinn2023reflexion}.
Hybrid paradigms combine symbolic planners utilizing code or graph \citep{liu2023llm+, cao-etal-2025-pgpo, zhang2025plan}, further improve generalization.  
\textbf{Multi-agent system planning.}
Multi-agent systems (MAS) exhibit great progress in complex tasks, where agents with diverse profiles collaborate, where planning becomes even more pronounced.
Early frameworks \citep{wu2023autogenenablingnextgenllm, hong2024metagpt} established structured workflows under centralized control, where a meta-agent is responsible for inter-agent planning, like role assignment and coordination. 
Recently, flexible planning strategies encourage hierarchical and partially decentralized planning \cite{zhang2025agentorchestra, li2025agentoriented}.

\subsection{Parallelable Agent}

Recent efforts have explored making LLM-based agents parallelizable to enhance efficiency.
Adaptive reasoning frameworks enable concurrent thought processes \citep{pan2025learning, pmlr-v235-zhang24au}, while DAG-based and divide–aggregate methods support simultaneous tool use \citep{zhang2025ecoact,zhu2025divide}.
At the system level, asynchronous planning–acting architectures and graph-based schedulers \citep{zhang2025plan} facilitate concurrent execution and coordination.
Despite these advances, research remains disparate across reasoning and scheduling, with no unified benchmark for evaluating agents' ability to exploit parallelism effectively.

\subsection{Benchmarks for Agent Planning}

\begin{table}[t]
\centering
\setlength{\belowcaptionskip}{-0.5cm}
\small
\resizebox{\linewidth}{!}{
\begin{tabular}{p{2.5cm}p{0.5cm}p{0.8cm}p{0.7cm}p{0.7cm}p{0.6cm}p{0.6cm}}
\toprule
\textbf{Benchmark} & \textbf{MAS} & \textbf{Interact.} & \textbf{IntraP.} & \textbf{InterP.} & \textbf{Time} & \textbf{Step} \\
\midrule
AsyncHow & \cmark & \xmark & \cmark & \cmark & \cmark & Short \\
TimeArena & \xmark & \xmark & \cmark & \xmark & \cmark & Short \\
Robotouille & \xmark & \cmark & \cmark & \xmark & \cmark & Long \\
WORFBENCH & \cmark & \xmark & \xmark & \cmark & \xmark & Short \\
Collab-Overcooked & \cmark & \cmark & \xmark & \cmark & \xmark & Long \\
CookBench & \xmark & \cmark & \cmark & \xmark & \xmark & Long \\
Overcooked-AI & \cmark & \cmark & \cmark & \cmark & \xmark & Long \\
\midrule
\textbf{ParaCook (Ours)} & \cmark & \cmark & \cmark & \cmark & \cmark & Long \\
\bottomrule
\end{tabular}
}
\caption{Comparison of agent planning benchmarks. IntraP: Intra-Agent Parallelism; InterP: Inter-Agent Parallelism; Env. Interact.: Environment Interaction; Time Eval.: Time Efficiency Evaluation as primary metric.}
\label{tab:benchmark_comparison}
\end{table}

Rigorous benchmarks are essential for systematic evaluation, especially in multi-agent settings. 
However, most existing benchmarks are not designed to evaluate time-efficient parallel scheduling under execution time constraints, and therefore cannot assess how effectively agents exploit parallelism to minimize overall completion time.
As summarized in Table \ref{tab:benchmark_comparison}, none of these benchmarks jointly evaluate intra- and inter-agent parallelism, and time efficiency under realistic execution dynamics. ParaCook is designed to fill this gap by explicitly modeling execution time, asynchronous waiting, and shared-resource contention in a multi-agent embodied environment. Detailed comparisons with benchmarks are provided in Appendix \ref{sec:benchmark_comparisons}.

\section{Formulation}
\label{formu}
In this section, we formulate the research question of \paraplan{} for agent systems, introducing basic concepts and key properties.

\paragraph{Task Decomposition and Dependence} 
First, we define the \paraplan{} uniformly for single- and multi-agent systems.
A complex task $\mathcal{I}$ can be decomposed into subtasks with dependencies, formalized as a Directed Acyclic Graph (DAG) $\displaystyle \gG = (\mathcal{V}, \mathcal{E})$. 
The vertex set $\mathcal{V} = \{v_1, v_2, \ldots, v_n\}$ represents subtasks, while the edge set $\mathcal{E} = \{(u, v)\}$ represents dependencies, i.e., $u$ must complete before $v$ starts. 
Formally, the predecessors of $v$ is denoted as $\displaystyle \parents_\gG(\ervv) = \{u \mid (u, v) \in \mathcal{E}\}$.
In terms of a \mas{} with $m$ participants, $\mathcal{A} = \{a_1, a_2, \ldots, a_m\}$, each vertex includes an extra attribute of agent identity to indicate the operator of each subtask, $v_j.id \leftarrow a_j$.
To build agent systems that achieve tasks efficiently, our study concentrates on task success and time efficiency.
\paragraph{Task Success}
Task success requires that the task decomposition $\gG=(\mathcal{V},\mathcal{E})$ is valid, all dependency constraints are satisfied, and every subtask $v\in\mathcal{V}$ is correctly executed by its assigned agent $v.id$.

\paragraph{Time Efficiency}
In consideration of the time efficiency, each subtask has a \textit{theoretical execution time}, $v_j.time \leftarrow t(v)$.
Subtasks involve inherent execution delays that proceed automatically, requiring only waiting rather than continuous involvement.
This allows parallel execution for \sas{} on long-horizon tasks.
For example,  when boiling water, the heating process requires only waiting without further action.
These are denoted by a delay function $d(u, v)$, modeling the minimum waiting interval between completing $u$ and starting $v$. 
\mas{}'s parallel execution includes (i) single-agent parallelism as defined by $d()$ and (ii) inter-agent parallelism: different agents execute independent subtasks simultaneously as defined by $\displaystyle \parents_\gG()$.

To reflect the physical world, we account for the stochastic nature of execution times. 
Our notion of \textit{actual execution time}, $t^\prime(v)$ incorporates inter-task transitions.
For example, the time required for an agent to be prepared for the next subtask. Such setup time between subtasks depends on the agent’s assigned task sequence in the plan.


\begin{figure*}[t]
\centering
\setlength{\belowcaptionskip}{-0.5cm}
\includegraphics[width=1\linewidth]{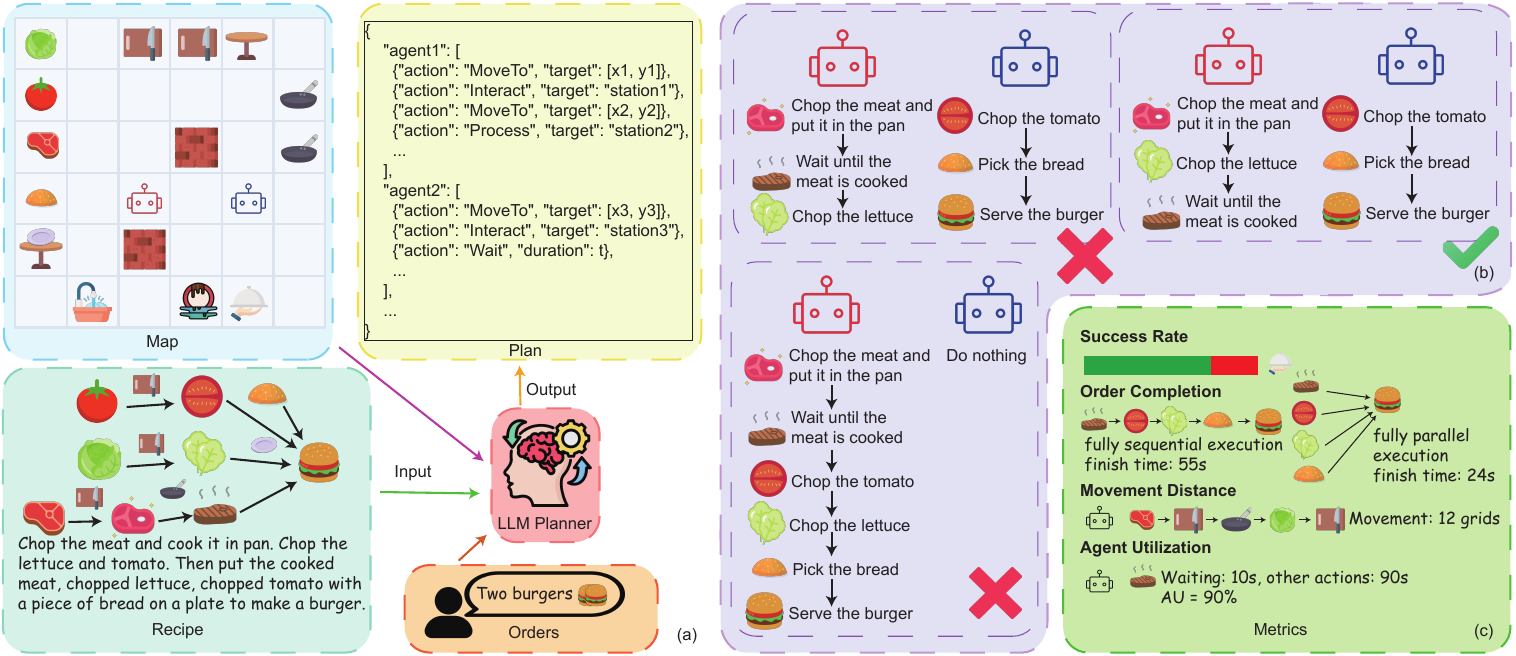}
\caption{Overview of the ParaCook benchmark, showing (a) the benchmark pipeline, (b) comparison of planning strategies, and (c) evaluation metrics for model performance.}
\label{fig:main_figure}
\end{figure*}

\section{Benchmark: ParaCook}

Based on our problem formulation, we propose ParaCook, the first benchmark that focuses on time-efficient \paraplan{} and multi-agent scheduling. 
We adopt ``cooking'' as a natural testbed, as it inherently exhibits sequential dependencies and asynchronous waiting, making it both challenging and well-suited for \paraplan{}.

\subsection{Environment}

The environment of ParaCook is a 2D grid world that simulates a kitchen, supporting agent –environment and agent–agent interactions while providing feedback. At the beginning of an episode, the environment initializes an agent team, the task to be performed, and the map configuration. 


\textbf{Action space} supported by our environment are
\begin{itemize}
    \item \textbf{MoveTo.} Specify a target coordinate for the agent to move to. Our environment computes the shortest path and executes the movement. The movement time is proportional to the distance.
    \item  \textbf{Interact.} Pick up or put down an item at the current location. The execution time is negligible.
    \item  \textbf{Process.} Perform a continuous operation at a workstation, such as chopping. The execution time depends on the specific operation.
    \item \textbf{Wait.} 
    Do nothing for a specified period, typically when blocked or waiting for a predecessor.
    \item \textbf{Finish.} 
    Declare task accomplished.
\end{itemize}
To isolate parallel scheduling from path-finding complexity, agents can overlap during movement without collisions, and item transfers occur via interaction with surfaces rather than direct handoffs.
Action execution times are fixed and summarized in Appendix \ref{sec:consts}.

\paragraph{Instantiation}
Formulated concepts in Section \ref{formu} are instantiated in our cooking environment.
\begin{itemize}
    \item Task decomposition $\displaystyle \gG = (\mathcal{V}, \mathcal{E})$ models the workflow of multiple orders, where nodes correspond to individual cooking steps and edges indicate dependency relations among them.
    \item Time delay $d()$ corresponds to cooking and waiting times between steps.
    \item Parallel execution enables simultaneous cooking operations and cross-dish coordination.
    \item Theoretical execution time $t()$ 
    \item Actual execution time $t^\prime()$ incorporates the travel time required for an agent to move to the next workstation for the following subtask. 
\end{itemize}

\subsection{Task}

The \as{} is instructed to propose action sequences to complete all dishes, given the input specified order.
We structure tasks based on recipes and orders, creating controllable yet meaningful concurrency challenges.

\paragraph{Recipes} The recipes describe the steps required to prepare a dish in natural language. For example, \textit{Put chopped lettuce and chopped tomato together on a plate to make a salad.} 
Our recipes vary in planning difficulty.
\begin{itemize}
    \item Simple recipes require only basic operations such as chopping and plating, providing a baseline for sequential planning.
    \item Intermediate recipes involve multiple operations (e.g., cut then cook) with asynchronous waiting periods, requiring agents to schedule efficiently.
    \item Complex recipes contain a larger number of ingredients that each require separate processing, demanding long-horizon planning and careful coordination to minimize idle time.
\end{itemize}


\paragraph{Orders} 
The orders are combinations of dish sequences. Orders with more dishes are more difficult, as the action horizon increases while conflicts and constraints of resources arise, including plate reuse and workstation availability. 
Thus, effective planning requires understanding and coordinating the steps of each dish to maximize time efficiency. 

Therefore, our task design enables concurrency (i) between steps within a dish, different steps can be executed simultaneously, and (ii) across dishes, idle time slots of earlier dishes can be used to prepare for later dishes.

\subsection{Map}
The configuration of the environment maps enables dynamic and flexible control over map size, workstation arrangement, agent count, and initial positions, ensuring high scalability.
More importantly, to support efficiency-oriented evaluation, the map can be configured by tuning the balance between workstations and agents, enabling controlled parallelism that ensures sufficient concurrent executions without excessive parallelization.

\subsection{Difficulty Control}
Based on the diverse and complex settings introduced above, the difficulty $D$ of a task can be formalized as
{\small
\setlength{\abovedisplayskip}{3pt}
\setlength{\belowdisplayskip}{1pt}
\begin{equation}
D = f(C_{\text{recipe}}, C_{\text{order}}, C_{\text{map}}),
\end{equation}
}
where $C_{\text{recipe}}$ denotes the operating complexity of selected dishes, $C_{\text{order}}$ denotes the composition order of dishes. $C_{\text{map}}$ indicates the challenges in spatial map configuration.
Therefore, ParaCook enables fine-grained difficulty control, facilitating systematic benchmarking of planning capabilities.

\section{Metrics}

ParaCook evaluates plans primarily along \emph{correctness} and \emph{efficiency}.  
This section details \textbf{Success Rate } for correctness, \textbf{Order Completion Time} for efficiency, and auxiliary metrics including \textbf{Movement Distance} and \textbf{Agent Utilization}.

\subsection{Primary Metrics}

\paragraph{Success Rate (SR)}  
At the dataset level, we define a run as successful if all dishes in the given order are completed correctly:
{\small
\setlength{\abovedisplayskip}{3pt}
\setlength{\belowdisplayskip}{-0.5pt}
\begin{equation}
SR = N_{\text{success}}/N_{\text{total}},
\end{equation}
}
where $N_{\text{success}}$ is the number of successful runs and $N_{\text{total}}$ is the total number of test cases.

\paragraph{Order Completion Time (OCT)}  
For a single successful task, OCT is the total elapsed time until completion of all orders, which the real time returned by the environment simulation.
{\small
\setlength{\abovedisplayskip}{3pt}
\setlength{\belowdisplayskip}{3pt}
\begin{equation}
OCT = T_{\text{actual}}.
\end{equation}
}
To aggregate across samples, we use two variants:
{\small
\setlength{\abovedisplayskip}{3pt}
\setlength{\belowdisplayskip}{-0.5pt}
\begin{equation}
pOCT = 1/N_{\text{total}} \sum_{i=1}^{N_{\text{total}}} T_i^\ast,
\end{equation}
}
{\small
\setlength{\abovedisplayskip}{3pt}
\setlength{\belowdisplayskip}{-0.5pt}
\begin{equation}
nOCT = 1/N_{\text{success}} \sum_{i=1}^{N_{\text{success}}} OCT_i / T_{\max}^i,
\end{equation}
}

where $T_i^\ast = OCT_i$ if task $i$ succeeds, and $T_i^\ast = T_{\max}^i$ otherwise.  
Here, $T_{\max}^i$ is the predefined upper bound for task $i$, details are provided in Appendix \ref{sec:upperbound}.  
$pOCT$ penalizes failures with maximal time, while $nOCT$ normalizes efficiency among successful runs.

\subsection{Auxiliary Metrics}

\paragraph{Movement Distance (MD)}  
For a task, MD is the mean travel distance of all agents:
{\small
\setlength{\abovedisplayskip}{3pt}
\setlength{\belowdisplayskip}{-0.5pt}
\begin{equation}
MD = 1/M \sum_{j=1}^{M} d_j,
\end{equation}
}
where $d_j$ is the distance traveled by agent $j$, and $M$ is the number of agents.  
Similarly, to aggregate across tasks, we calculate the penalized form:
{\small
\setlength{\abovedisplayskip}{3pt}
\setlength{\belowdisplayskip}{-0.5pt}
\begin{equation}
pMD = 1/N_{\text{total}} \sum_{i=1}^{N_{\text{total}}} D_i^\ast,
\end{equation}
}
with $D_i^\ast = MD_i$ for successful tasks and $D_i^\ast = D_{\max}^i$ otherwise. $D_{\max}^i$ represents the predefined upper bound of movement distance for task $i$, details are provided in Appendix \ref{sec:upperbound}.

\paragraph{Agent Utilization (AU)}  
For each agent $j$, utilization is defined as the proportion of active working time:
{\small
\setlength{\abovedisplayskip}{3pt}
\setlength{\belowdisplayskip}{-0.5pt}
\begin{equation}
u_j = T^{\text{work}}_j/T_j^{\text{total}},\\
AU = 1/M \sum_{j=1}^{M} u_j.
\end{equation}
}
Dataset-level AU is averaged over successful runs.

In summary, \textbf{SR} evaluates correctness, and \textbf{OCT} evaluates efficiency with penalized and normalized variants.  
\textbf{MD} and \textbf{AU} serve as auxiliary analyses, capturing execution cost and coordination quality.  

\section{Experiments}
This section presents our empirical implementations and main findings from our results.
\subsection{Experimental Setup}
\paragraph{Dataset} We manually annotated six categories of recipes with increasing levels of difficulty, and each category contains several different dishes. For orders, a set of dishes was randomly sampled from a given recipe category, with the number of dishes ranging from one to four to control the complexity of the task. To encourage parallelism, each kitchen map was equipped with two stations of each type, placed randomly while ensuring connectivity. The number of agents varied between one and three. For each configuration, we used five random seeds to generate different orders and maps. All evaluation configurations and instance counts are enumerated in Appendix \ref{sec:eval_config}.

\paragraph{Agents} 
We evaluate different centralized planning methods, where a single LLM planner generates actions for all agents. We consider the I/O and CoT prompting methods. I/O produces a complete plan directly from the initial state and specifies the actions for each agent. 
CoT \citep{wei2022chain} encourages reasoning while planning, explicitly predicting intermediate states. 
We evaluate state-of-the-art API LLMs, including GPT-5, Claude-Opus-4.1, Gemini-2.5-Pro \cite{comanici2025gemini25pushingfrontier}, DeepSeek-V3.2-Exp \cite{deepseekai2024deepseekv32}, and Qwen3-Max-Preview.

\begin{table*}[t]
\centering
\setlength{\belowcaptionskip}{-0.6cm}
\small
\setlength{\tabcolsep}{3pt}
\resizebox{\textwidth}{!}{
\begin{tabular}{l rrrr rrrr rrrr}
\toprule
& \multicolumn{4}{c}{Success Rate (SR, \%)} & \multicolumn{4}{c}{Penalized OCT (pOCT)} & \multicolumn{4}{c}{Normalized OCT (nOCT)} \\
\cmidrule(lr){2-5} \cmidrule(lr){6-9} \cmidrule(lr){10-13}
Model & Easy & Medium & Hard & Avg. & Easy & Medium & Hard & Avg. & Easy & Medium & Hard & Avg. \\
\midrule
GPT-5             & \textbf{80.83}{\scriptsize $\pm$7} & \textbf{69.17}{\scriptsize $\pm$8} & \textbf{45.00}{\scriptsize $\pm$9} & \textbf{65.00} & \textbf{137.98}{\scriptsize $\pm$26} & \textbf{330.80}{\scriptsize $\pm$60} & \textbf{416.35}{\scriptsize $\pm$57} & \textbf{295.04} & \textbf{29.16}{\scriptsize $\pm$2} & \textbf{25.38}{\scriptsize $\pm$2} & \textbf{27.34}{\scriptsize $\pm$2} & \textbf{27.29} \\
Gemini-2.5-Pro    & 60.00{\scriptsize $\pm$9} & 55.07{\scriptsize $\pm$12} & 27.14{\scriptsize $\pm$10} & 47.40$_{\textcolor{red}{-17.60}}$ & 191.94{\scriptsize $\pm$32} & 390.33{\scriptsize $\pm$83} & 486.94{\scriptsize $\pm$73} & 356.40$_{\textcolor{red}{+61.36}}$ & 31.24{\scriptsize $\pm$2} & 27.68{\scriptsize $\pm$2} & 30.56{\scriptsize $\pm$4} & 29.83$_{\textcolor{red}{+2.54}}$ \\
DeepSeek-V3.2-Exp & 66.67{\scriptsize $\pm$8} & 47.50{\scriptsize $\pm$9} & 21.67{\scriptsize $\pm$7} & 45.28$_{\textcolor{red}{-19.72}}$ & 176.89{\scriptsize $\pm$30} & 467.98{\scriptsize $\pm$68} & 521.19{\scriptsize $\pm$59} & 388.69$_{\textcolor{red}{+93.65}}$ & 30.01{\scriptsize $\pm$2} & 26.31{\scriptsize $\pm$2} & 26.44{\scriptsize $\pm$4} & 27.59$_{\textcolor{red}{+0.30}}$ \\
Claude-Opus-4.1   & 26.67{\scriptsize $\pm$8} & 12.50{\scriptsize $\pm$9} & 0.00{\scriptsize $\pm$0}  & 13.06$_{\textcolor{red}{-51.94}}$ & 268.54{\scriptsize $\pm$34} & 585.65{\scriptsize $\pm$93} & --     & -- & 31.96{\scriptsize $\pm$3} & 35.64{\scriptsize $\pm$5} & -- & -- \\
Qwen3-Max-Preview & 6.67{\scriptsize $\pm$4}  & 0.00{\scriptsize $\pm$0}  & 0.00{\scriptsize $\pm$0}  & 2.22$_{\textcolor{red}{-62.78}}$  & 291.97{\scriptsize $\pm$31} & --     & --     & -- & 31.33{\scriptsize $\pm$5} & -- & -- & -- \\
\bottomrule
\end{tabular}
}
\caption{Success Rate (SR), Penalized and Normalized Order Completion Time (pOCT, nOCT) of different models across three difficulty levels and averages. Values shown as mean $\pm$ 95\% confidence interval. \textbf{Bold} indicates the best performance. Red subscripts in the Avg. column show the performance gap relative to the best model (GPT-5).}
\label{tab:main_results_sr_poct_avg}
\end{table*}

\subsection{Results}

Table \ref{tab:main_results_sr_poct_avg} presents overall results, while Table \ref{tab:main_results} shows more detailed scores. 
Our results answer the following three key questions.

\paragraph{ParaCook is challenging enough to differentiate SOTA LLMs, which vary on SR.}
In terms of \textit{Success Rate (SR)}, GPT-5 achieves the highest success rate across all difficulty levels, with an average SR of 65.0\%. It maintains strong performance on Easy (80.8\%) and Medium (69.2\%) tasks, and remains competitive even on Hard tasks (45.0\%). In contrast, Gemini-2.5-Pro (47.4\% on average) and DeepSeek-V3.2-Exp (45.3\%) form the second tier, showing moderate performance but a noticeable drop in the Hard setting (27.1\% and 21.7\%, respectively). Claude-Opus-4.1 (13.1\%) and Qwen3 (2.2\%) perform poorly, failing almost completely on Medium and Hard tasks.



\paragraph{ParaCook serves as a suitable, still challenging testbed for time efficiency for planning capabilities of SOTA LLMs.}
In terms of pOCT, GPT-5 demonstrates the best performance, with the lowest average planning overhead (295.0), consistently faster than other models at all difficulty levels. Gemini-2.5-Pro (356.4) and DeepSeek-V3.2-Exp (388.7) occupy the middle range, showing reasonable efficiency but still substantially slower than GPT-5. Examining efficiency on successful runs (nOCT), GPT-5 achieves the best average of 27.29, with DeepSeek-V3.2-Exp close behind at 27.59, while Gemini-2.5-Pro lags at 29.83. This indicates that even among successful plans, GPT-5 consistently produces more time-efficient schedules. Claude-Opus-4.1 and Qwen3 exhibit unstable performance: despite occasionally achieving low overhead on Easy tasks, they either fail to complete tasks at higher difficulty levels or produce unreliable results, limiting their practical applicability.

In detail, Table \ref{tab:main_results} illustrates that time efficiency is also reflected in MD and AU. Specifically, GPT-5 maintains minimal pMD across all tasks and high agent utilization (AU) above 86\%. Gemini-2.5-Pro and DeepSeek-V3.2-Exp are less efficient, with pMD values rising to 178.60 and 195.28 on Medium tasks, respectively. Claude-Opus-4.1 performs the worst, with pMD exceeding 230 and significantly lower utilization, confirming that inefficient scheduling and the failure to parallelize actions directly lead to longer completion times.

\paragraph{SOTA LLM planners demonstrably lag behind human performance in success rate, time efficiency, and spatial optimization.}
We provide human performance to answer the question: \textit{How do LLM planners compare to human performance?}
\begin{figure}[ht]
    \centering
    \setlength{\belowcaptionskip}{-0.4cm}
    \setlength{\abovecaptionskip}{0.2cm}
    \includegraphics[width=\linewidth]{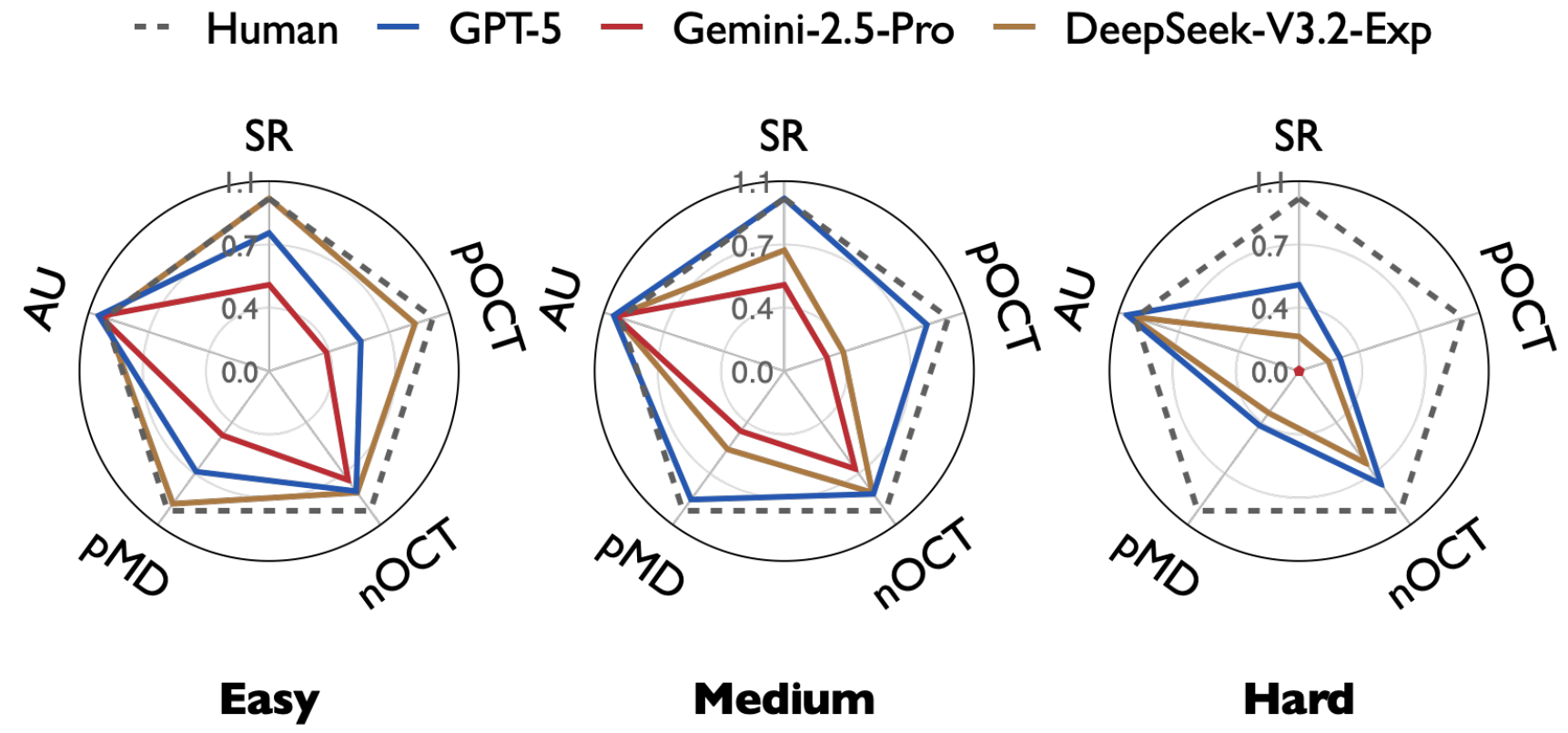} 
    \caption{Human-evaluated subset for LLM-human comparison. Detailed scores are in Table \ref{tab:human_subset_results} and \ref{tab:human_subset_results_normalized_per_difficulty}.}
    \label{fig:human}
\end{figure}
Figure \ref{fig:human} compares human participants with top-performing LLMs on a curated subset of tasks (whose detailed results are presented in Table \ref{tab:human_subset_results}). Details of the human evaluation protocol and interface are provided in Appendix \ref{sec:human_gui}. Humans achieve a perfect 100\% success rate across all difficulty levels, highlighting their robustness in complex, long-horizon scenarios where LLMs may fail. The performance gap becomes particularly evident on Hard tasks. For instance, the top-performing model, GPT-5 with CoT, achieved a 57\% SR, whereas humans made no errors.

In terms of efficiency, while top LLMs are competitive on easy tasks, they lag significantly as complexity increases. On Hard tasks, humans completed orders with an nOCT of 15.38, whereas GPT-5 required a longer time of 17.61. The disparity in spatial efficiency is even more stark, with humans having a pMD of just 47.75, compared to GPT-5's 143.55, indicating far more optimized movement from humans. Interestingly, AU is slightly lower for humans (87.13\%) than for GPT-5 (94.34\%). This is not due to inefficiency, but because human plans are executed so quickly that the necessary waiting time between dependent tasks constitutes a larger proportion of the total duration. Overall, these results demonstrate that while LLMs can produce correct and parallelized plans for less complex tasks, their execution strategies remain less optimized than those of humans, particularly for high coordination demands. A detailed time-budget decomposition is provided in Appendix \ref{sec:time_breakdown}.

\section{Analysis}

\begin{figure*}[t]
\setlength{\belowcaptionskip}{-0.3cm}
\setlength{\abovecaptionskip}{0.2cm}
\centering
\includegraphics[width=1\linewidth]{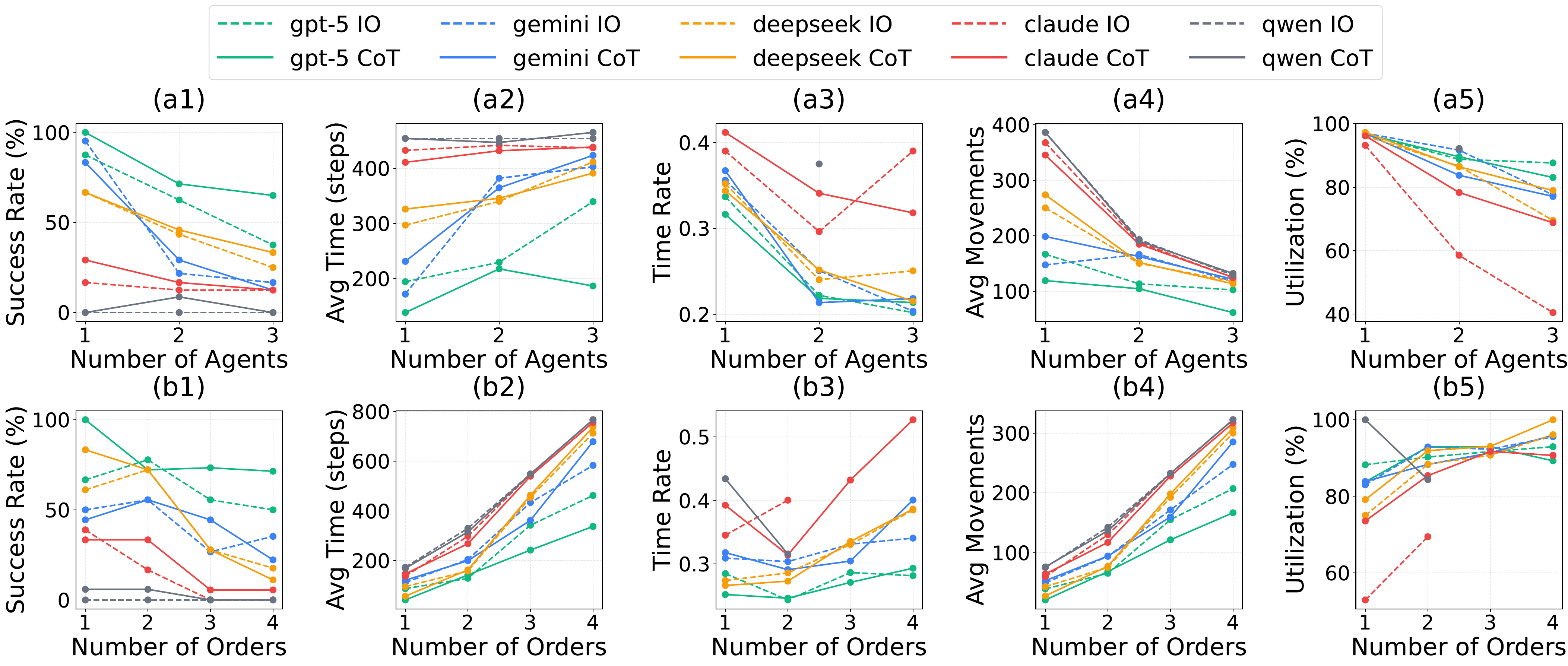}
\vspace{-1mm}
\caption{Model performance across different task complexities. Top row (a1-a5): varying number of agents. Bottom row (b1-b5): varying number of orders. Each column corresponds to the five metrics.}
\label{fig:agent_order_num_lines}
\end{figure*}

\subsection{Performance across Task Complexity}

Figure \ref{fig:agent_order_num_lines} shows the performance by varying two key complexity dimensions: the number of agents and the number of orders.

\paragraph{Impact of Agent Number: A team with more agents requires better \paraplan{}, but increasing coordination complexity and resource contention.}
As shown in Figure (a1), the SR consistently declines as the number of agents increases, highlighting the inherent difficulty of multi-agent coordination. This is also reflected in the pOCT (Figure (a2)), which generally rises with the failure rate. A notable exception is GPT-5 CoT, whose pOCT drops at three agents, suggesting its successful plans achieve exceptional parallel efficiency that outweighs the penalty from a slightly lower SR. The nOCT offers deeper insight (Figure (a3)). The sharp drop from one to two agents demonstrates that an additional agent effectively enhances efficiency through parallelism. However, this gain diminishes from two to three agents, where the nOCT curve flattens. We attribute this to resource contention; with a limited number of workstations (e.g., two cutting boards), a third agent cannot work in parallel on the same task type, capping the achievable speedup. Furthermore, the anomalous rise in Claude IO's nOCT, coupled with its significantly lower AU (Figure (a5)), suggests a "false parallelism" where one agent works while others wait, failing to leverage the multi-agent setup. Also, the expected decrease in pMD (Figure (a4)) confirms better task locality with more agents.

\paragraph{Impact of Order Number: Increasing order number raises cost and lowers success, but sometimes allows more parallelism opportunity.}
In terms of the planning horizon,
increasing the number of orders extends the task sequence, leading to a general decline in SR (Figure (b1)). Consequently, the overall cost metrics, pOCT (Figure (b2)) and pMD (Figure (b4)), rise steadily as more work requires more time and movement. The nOCT trend is more nuanced (Figure (b3)); the metric remains stable or even decreases when moving from one to two orders. This is because sufficient initial resources allow agents to prepare two orders in parallel, so the actual completion time $T_{\text{actual}}$ increases only modestly. In contrast, the normalization factor, $T_{\max}$, is set significantly higher for a two-order task. This combination results in a stable or smaller nOCT ratio, indicating scalable plan quality in scenarios with low complexity. Lastly, AU generally improves with more orders (Figure (b5)), as more subtasks effectively reduce agent idle time in successful runs.

\begin{table*}[t]
\centering
\setlength{\belowcaptionskip}{-0.5cm}
\setlength{\abovecaptionskip}{0.2cm}
\small
\setlength{\tabcolsep}{4pt}
\resizebox{\textwidth}{!}{
\begin{tabular}{l l ll ll ll ll }
\toprule
& & \multicolumn{2}{c}{Easy} & \multicolumn{2}{c}{Medium} & \multicolumn{2}{c}{Hard} & \multicolumn{2}{c}{Average} \\
\cmidrule(lr){3-4} \cmidrule(lr){5-6} \cmidrule(lr){7-8} \cmidrule(lr){9-10}
Model & Method & SR$\uparrow$ & pOCT$\downarrow$  & SR$\uparrow$ & pOCT$\downarrow$ & SR$\uparrow$ & pOCT$\downarrow$ & SR$\uparrow$ & pOCT$\downarrow$ \\
\midrule
\multirow{2}{*}{GPT-5} & IO    & 80.83{\scriptsize $\pm$7} & 137.98{\scriptsize $\pm$26}  & 69.17{\scriptsize $\pm$8} & 330.80{\scriptsize $\pm$60}  & 45.00{\scriptsize $\pm$9} & 416.35{\scriptsize $\pm$57} & 65.00 & 295.04 \\
                       & CoT   & 84.17{\scriptsize $\pm$7}$_{\textcolor{blue}{+3.34}}$ & 129.59{\scriptsize $\pm$27}$_{\textcolor{blue}{-8.39}}$  & 77.03{\scriptsize $\pm$10}$_{\textcolor{blue}{+7.86}}$ & 283.03{\scriptsize $\pm$73}$_{\textcolor{blue}{-47.77}}$ & 57.39{\scriptsize $\pm$9}$_{\textcolor{blue}{+12.39}}$ & 359.16{\scriptsize $\pm$60}$_{\textcolor{blue}{-57.19}}$ & 72.86$_{\textcolor{blue}{+7.86}}$ & 257.26$_{\textcolor{blue}{-37.78}}$ \\
\midrule
\multirow{2}{*}{Gemini-2.5-Pro} & IO    & 60.00{\scriptsize $\pm$9} & 191.94{\scriptsize $\pm$32} & 55.07{\scriptsize $\pm$12} & 390.33{\scriptsize $\pm$83}  & 27.14{\scriptsize $\pm$10} & 486.94{\scriptsize $\pm$73} & 47.40 & 356.40 \\
                                & CoT   & 55.83{\scriptsize $\pm$9}$_{\textcolor{red}{-4.17}}$ & 201.43{\scriptsize $\pm$31}$_{\textcolor{red}{+9.49}}$  & 47.22{\scriptsize $\pm$12}$_{\textcolor{red}{-7.85}}$ & 419.10{\scriptsize $\pm$82}$_{\textcolor{red}{+28.77}}$  & 37.50{\scriptsize $\pm$11}$_{\textcolor{blue}{+10.36}}$ & 447.49{\scriptsize $\pm$75}$_{\textcolor{blue}{-39.45}}$ & 46.85$_{\textcolor{red}{-0.55}}$ & 356.01$_{\textcolor{blue}{-0.39}}$ \\
\midrule
\multirow{2}{*}{DeepSeek-V3.2-Exp} & IO    & 66.67{\scriptsize $\pm$8} & 176.89{\scriptsize $\pm$30} & 47.50{\scriptsize $\pm$9} & 467.98{\scriptsize $\pm$68} & 21.67{\scriptsize $\pm$7} & 521.19{\scriptsize $\pm$59} & 45.28 & 388.69 \\
                                   & CoT   & 67.50{\scriptsize $\pm$8}$_{\textcolor{blue}{+0.83}}$ & 178.42{\scriptsize $\pm$32}$_{\textcolor{red}{+1.53}}$  & 45.83{\scriptsize $\pm$9}$_{\textcolor{red}{-1.67}}$ & 470.74{\scriptsize $\pm$70}$_{\textcolor{red}{+2.76}}$  & 40.83{\scriptsize $\pm$9}$_{\textcolor{blue}{+19.16}}$ & 465.84{\scriptsize $\pm$64}$_{\textcolor{blue}{-55.35}}$ & 51.39$_{\textcolor{blue}{+6.11}}$ & 371.67$_{\textcolor{blue}{-17.02}}$ \\
\midrule
\multirow{2}{*}{Claude-Opus-4.1} & IO    & 26.67{\scriptsize $\pm$8} & 268.54{\scriptsize $\pm$34} & 12.50{\scriptsize $\pm$9} & 585.65{\scriptsize $\pm$93}  & 0.00{\scriptsize $\pm$0} & --  & 13.06 & 427.10 \\
                                 & CoT   & 55.00{\scriptsize $\pm$9}$_{\textcolor{blue}{+28.33}}$ & 215.99{\scriptsize $\pm$34}$_{\textcolor{blue}{-52.55}}$  & 14.58{\scriptsize $\pm$10}$_{\textcolor{blue}{+2.08}}$ & 577.52{\scriptsize $\pm$94}$_{\textcolor{blue}{-8.13}}$ & 0.00{\scriptsize $\pm$0}$_{\textcolor{black}{+0.00}}$ & --  & 23.19$_{\textcolor{blue}{+10.13}}$ & 396.76$_{\textcolor{blue}{-30.34}}$ \\
\midrule
\multirow{2}{*}{Qwen3-Max-Preview} & IO    & 6.67{\scriptsize $\pm$4} & 291.97{\scriptsize $\pm$31} & 0.00{\scriptsize $\pm$0} & --  & 0.00{\scriptsize $\pm$0} & -- & 2.22 & -- \\
                                   & CoT   & 10.83{\scriptsize $\pm$6}$_{\textcolor{blue}{+4.16}}$ & 286.73{\scriptsize $\pm$31}$_{\textcolor{blue}{-5.24}}$  & 0.00{\scriptsize $\pm$0}$_{\textcolor{black}{+0.00}}$ & --  & 0.00{\scriptsize $\pm$0}$_{\textcolor{black}{+0.00}}$ & -- & 3.61$_{\textcolor{blue}{+1.39}}$ & --\\
\bottomrule
\end{tabular}
}
\caption{Results with CoT. Values are reported as mean $\pm$ 95\% confidence interval. Blue subscripts indicate improvements (higher SR or lower pOCT), red indicates degradation.}
\label{tab:main_results_cot}
\end{table*}

\subsection{Performance across Prompting and Planning Methods}
Does Chain-of-Thought prompting universally improve model performance? Our findings indicate that \textbf{the effectiveness of CoT is heavily dependent on the model's underlying reasoning capability.}

For a strong model like GPT-5, CoT acts as a consistent amplifier, enhancing performance across the board and raising the SR on Hard tasks from 45\% to 57\%. In contrast, for moderately capable models, CoT's influence is unstable. Gemini, for instance, showed a counterintuitive decline in success on Easy and Medium tasks when using CoT. DeepSeek presented a mixed picture, with CoT only providing a decisive improvement on Hard tasks, where the SR rose from 22\% to 41\%. For weaker models such as Claude and Qwen, CoT functioned as a limited rescue mechanism, significantly improving success on Easy tasks but remaining insufficient for Medium or Hard problems.
Therefore, CoT is not a universally beneficial strategy. It reliably enhances already strong models but can destabilize moderately capable ones and offers only limited support to weaker ones.

Beyond CoT, we also evaluated graph-augmented prompting and interactive planning. These methods show model-dependent effectiveness with no universally superior approach. Full results are provided in Appendix \ref{sec:advanced_methods}.

\subsection{Abstract Task}

\paragraph{Method}
To further investigate the essence of suboptimal temporal efficiency, we designed abstract planning tasks that isolate planning from action execution.
In this setup, tasks are formulated as in section \ref{formu}. The LLM’s objective is to generate a schedule, allocating and ordering subtasks for a given number of agents, that minimizes the total completion time. We benchmarked the generated plans against the provably optimal solutions derived from a constraint programming solver.
We analyze three aspects, which are Success Rate, nOCT, and pOCT, with a penalty of 1.2 times the optimal time to any invalid plan.

\paragraph{Results}
Table \ref{tab:abstract_results} shows results on abstract tasks, where we summarize two key findings.

\textbf{SOTA LLMs have strong inherent potential in reasoning for \paraplan{}.}
For instance, GPT-5, Gemini-2.5-Pro, and Claude-Opus all achieved perfect SR scores, with nOCT scores of 1.0174, 1.0161, and 1.0721 respectively, indicating their schedules were on average only 1-7\% slower than the theoretical optimum. This demonstrates a strong inherent capability for reasoning about complex temporal dependencies and optimizing for parallelism when not constrained by environmental interaction.

\textbf{The capability hierarchy observed in the ParaCook largely persists in the abstract task.} 
For example, Qwen3-Max-Preview’s performance is consistently inferior, whose SR falls to 79\% and nOCT shows nearly 13\% less efficient on average.
This suggests that the performance differences among LLMs are at least partially rooted in their fundamental optimization and reasoning abilities, not solely in their capacity to handle the specific rules of the embodied environment. 

Crucially, the contrast between near-optimal performance on abstract tasks and the inefficiency in real ParaCook highlights the need for structured approaches as a bridge, such as hierarchical planning frameworks that separate high-level scheduling from detailed action execution. However, while abstract schedules provide optimal solutions, extending such methods to embodied ParaCook tasks is non-trivial, as the intricate spatial-temporal constraints make a complete symbolic abstraction of the environment extremely difficult, see Appendix \ref{sec:or_discussion} for a detailed discussion.

\begin{table}[h]
\centering
\small
\setlength{\belowcaptionskip}{-0.8cm}
\setlength{\abovecaptionskip}{0.2cm}
\begin{tabular}{l ccc}
\toprule
Model & SR$\uparrow$ & pOCT$\downarrow$ & nOCT$\downarrow$ \\
\midrule
cp\_sat & 100.00 & 100.28 & 1.0000 \\
GPT-5 & 100.00 & 103.76 & 1.0241 \\
Gemini-2.5-Pro & 100.00 & 103.35 & 1.0220 \\
DeepSeek-Reasoner & 94.00 & 110.11 & 1.0619 \\
Claude-Opus-4.1 & 100.00 & 112.87 & 1.0902 \\
Qwen3-Max-Preview & 78.00 & 117.96 & 1.1315 \\
\bottomrule
\end{tabular}
\caption{Model performance on abstract tasks.}
\label{tab:abstract_results}
\end{table}

\section{Conclusion}

In this work, we introduce ParaCook, a benchmark for evaluating time-efficient planning in multi-agent systems. We systematically test state-of-the-art LLMs across varied task complexities and find that GPT-5 achieves the best overall performance in both success rate and completion time. Human–LLM comparisons reveal that, while top models approach human performance in simple scenarios, they still lag in complex coordination and fine-grained temporal optimization. Results on abstract tasks further confirm LLMs’ strong reasoning ability for high-level scheduling. ParaCook establishes a scalable foundation for advancing time-efficient and coordination-aware LLM agents.

\section*{Limitations}

We acknowledge the limitations of this work. (i) Our current work mainly focuses on establishing a benchmark and evaluating existing methods, and does not propose a novel solution to overcome the parallel planning challenges. (ii) The scope of Agent methods tested was limited, primarily focusing on IO and CoT strategies. Our future work will focus on these two directions.

\bibliography{custom}

\appendix

\clearpage
\section{Benchmark-Specific Comparisons}
\label{sec:benchmark_comparisons}

This appendix provides brief, benchmark-specific comparisons highlighting how ParaCook differs from representative agent planning benchmarks along execution grounding, time efficiency, and parallelism.

\paragraph{AsyncHow} \citep{lin2024graphenhancedlargelanguagemodels} formulates scheduling as ordering oracle-decomposed subtasks with given durations and dependency graphs, reducing planning to dependency resolution rather than execution-grounded scheduling.
\paragraph{TimeArena} \citep{zhang-etal-2024-timearena} studies time-aware multitasking in a textual, single-agent simulation, without embodied execution, spatial constraints, or multi-agent coordination.
\paragraph{Robotouille} \citep{gonzalez-pumariega2025robotouille} simulates cooking tasks with proper time consumption, but only focuses on single-agent asynchronous planning. 
\paragraph{WORFBENCH} \citep{qiao2024benchmarking} decomposes tasks into workflow graphs but evaluates graph similarity rather than execution time. 
\paragraph{Collab-Overcooked} \citep{sun2025collabovercookedbenchmarkingevaluatinglarge} and \textbf{CookBench} \citep{cai2025cookbench} provide comprehensive challenges for MAS. However, they also do not support explicit time efficiency evaluation. 
\paragraph{Overcooked-AI} \citep{NEURIPS2019_f5b1b89d} evaluates human-AI coordination and adaptation to human behavior, rather than time-efficient parallel scheduling.

Overall, existing benchmarks address complementary aspects of agent planning but do not jointly evaluate execution-grounded time efficiency, intra-agent parallelism, and inter-agent parallelism in multi-agent settings, which ParaCook is specifically designed to assess.

\section{detailed experimental results}

Table \ref{tab:main_results}, \ref{tab:human_subset_results} and \ref{tab:human_subset_results_normalized_per_difficulty} provide detailed experimental results.

\begin{table*}[t]
\centering
\small
\resizebox{\textwidth}{!}{
\begin{tabular}{l l ccccc ccccc ccccc}
\toprule
& & \multicolumn{5}{c}{Easy} & \multicolumn{5}{c}{Medium} & \multicolumn{5}{c}{Hard} \\
\cmidrule(lr){3-7} \cmidrule(lr){8-12} \cmidrule(lr){13-17}
Model & Method & SR$\uparrow$ & pOCT$\downarrow$ & nOCT$\downarrow$ & pMD$\downarrow$ & AU$\uparrow$ & SR$\uparrow$ & pOCT$\downarrow$ & nOCT$\downarrow$ & pMD$\downarrow$ & AU$\uparrow$ & SR$\uparrow$ & pOCT$\downarrow$ & nOCT$\downarrow$ & pMD$\downarrow$ & AU$\uparrow$ \\
\midrule
\multirow{2}{*}{GPT-5} & IO    & 80.83 & 137.98 & 29.16 & 67.84 & 90.16 & 69.17 & 330.80 & 25.38 & 144.21 & 83.75 & 45.00 & 416.35 & 27.34 & 185.76 & 94.70 \\
                       & CoT   & 84.17 & 129.59 & 26.84 & 64.34 & 91.10 & 77.03 & 283.03 & 24.92 & 129.32 & 86.52 & 57.39 & 359.16 & 23.56 & 164.00 & 91.98 \\
\midrule
\multirow{2}{*}{Gemini-2.5-Pro} & IO    & 60.00 & 191.94 & 31.24 & 90.03 & 92.73 & 55.07 & 390.33 & 27.68 & 164.62 & 84.23 & 27.14 & 486.94 & 30.56 & 210.18 & 87.71 \\
                                & CoT   & 55.83 & 201.43 & 32.06 & 93.33 & 93.69 & 47.22 & 419.10 & 27.05 & 178.60 & 87.05 & 37.50 & 447.49 & 30.38 & 198.51 & 89.91 \\
\midrule
\multirow{2}{*}{DeepSeek-V3.2-Exp} & IO    & 66.67 & 176.89 & 30.01 & 83.49 & 89.52 & 47.50 & 467.98 & 26.31 & 193.85 & 84.89 & 21.67 & 521.19 & 26.44 & 229.07 & 95.43 \\
                                   & CoT   & 67.50 & 178.42 & 29.29 & 83.63 & 90.52 & 45.83 & 470.74 & 25.16 & 195.28 & 81.89 & 40.83 & 465.84 & 26.08 & 206.53 & 94.52 \\
\midrule
\multirow{2}{*}{Claude-Opus-4.1} & IO    & 26.67 & 268.54 & 31.96 & 122.12 & 70.95 & 12.50 & 585.65 & 35.64 & 236.99 & 52.33 & 0.00 & -- & -- & -- & -- \\
                                 & CoT   & 55.00 & 215.99 & 33.20 & 98.98 & 83.87 & 14.58 & 577.52 & 37.61 & 233.45 & 66.69 & 0.00 & -- & -- & -- & -- \\
\midrule
\multirow{2}{*}{Qwen3-Max-Preview} & IO    & 6.67 & 291.97 & 31.33 & 132.63 & 61.54 & 0.00 & -- & -- & -- & -- & 0.00 & -- & -- & -- & -- \\
                                   & CoT   & 10.83 & 286.73 & 33.55 & 130.47 & 90.27 & 0.00 & -- & -- & -- & -- & 0.00 & -- & -- & -- & -- \\
\bottomrule
\end{tabular}
}
\caption{Results of different methods on state-of-the-art models across three difficulty levels.}
\label{tab:main_results}
\end{table*}

\begin{table*}[t]
\centering
\small
\resizebox{\textwidth}{!}{
\begin{tabular}{l l ccccc ccccc ccccc}
\toprule
& & \multicolumn{5}{c}{Easy} & \multicolumn{5}{c}{Medium} & \multicolumn{5}{c}{Hard} \\
\cmidrule(lr){3-7} \cmidrule(lr){8-12} \cmidrule(lr){13-17}
Model & Method & SR$\uparrow$ & pOCT$\downarrow$ & nOCT$\downarrow$ & pMD$\downarrow$ & AU$\uparrow$ & SR$\uparrow$ & pOCT$\downarrow$ & nOCT$\downarrow$ & pMD$\downarrow$ & AU$\uparrow$ & SR$\uparrow$ & pOCT$\downarrow$ & nOCT$\downarrow$ & pMD$\downarrow$ & AU$\uparrow$ \\
\midrule
Human & -- & 100.00 & 44.00 & 19.48 & 32.17 & 93.31 & 100.00 & 73.67 & 16.24 & 57.17 & 87.96 & 100.00 & 66.67 & 15.31 & 51.08 & 90.92 \\
\midrule
\multirow{2}{*}{GPT-5} & IO & 80.00 & 78.60 & 22.70 & 44.60 & 97.09 & 100.00 & 85.20 & 18.53 & 62.15 & 91.60 & 50.00 & 270.50 & 19.00 & 130.55 & 95.67 \\
& CoT & 90.00 & 53.20 & 19.08 & 36.05 & 96.35 & 100.00 & 91.50 & 20.36 & 59.58 & 88.54 & 40.00 & 304.20 & 17.61 & 143.55 & 94.34 \\
\midrule
\multirow{2}{*}{Gemini-2.5-Pro} & IO & 50.00 & 126.90 & 24.84 & 69.70 & 96.46 & 50.00 & 288.33 & 23.19 & 132.58 & 90.66 & 0.00 & -- & -- & -- & -- \\
& CoT & 60.00 & 117.70 & 21.71 & 64.40 & 94.09 & 50.00 & 278.17 & 21.31 & 127.08 & 86.30 & 50.00 & 247.50 & 18.26 & 125.75 & 93.58 \\
\midrule
\multirow{2}{*}{DeepSeek-V3.2-Exp} & IO & 100.00 & 51.80 & 23.19 & 34.35 & 93.39 & 70.00 & 204.10 & 18.77 & 102.15 & 90.61 & 20.00 & 379.40 & 23.36 & 171.75 & 92.95 \\
& CoT & 100.00 & 49.30 & 22.32 & 33.85 & 93.63 & 90.00 & 137.80 & 21.54 & 76.80 & 90.17 & 50.00 & 263.70 & 18.90 & 130.50 & 99.78 \\
\bottomrule
\end{tabular}
}
\caption{Results on the human-evaluated subset for LLM-human comparison.}
\label{tab:human_subset_results}
\end{table*}

\begin{table*}[t]
\centering
\small
\resizebox{\textwidth}{!}{
\begin{tabular}{l l ccccc ccccc ccccc}
\toprule
& & \multicolumn{5}{c}{Easy} & \multicolumn{5}{c}{Medium} & \multicolumn{5}{c}{Hard} \\
\cmidrule(lr){3-7} \cmidrule(lr){8-12} \cmidrule(lr){13-17}
Model & Method & SR$\uparrow$ & pOCT$\uparrow$ & nOCT$\uparrow$ & pMD$\uparrow$ & AU$\uparrow$ & SR$\uparrow$ & pOCT$\uparrow$ & nOCT$\uparrow$ & pMD$\uparrow$ & AU$\uparrow$ & SR$\uparrow$ & pOCT$\uparrow$ & nOCT$\uparrow$ & pMD$\uparrow$ & AU$\uparrow$ \\
\midrule
Human & -- & 1.00 & 1.00 & 1.00 & 1.00 & 1.00 & 1.00 & 1.00 & 1.00 & 1.00 & 1.00 & 1.00 & 1.00 & 1.00 & 1.00 & 1.00 \\
\midrule
\multirow{2}{*}{GPT-5} & IO & 0.80 & 0.56 & 0.86 & 0.72 & 1.04 & 1.00 & 0.86 & 0.88 & 0.92 & 1.04 & 0.50 & 0.25 & 0.81 & 0.39 & 1.05 \\
& CoT & 0.90 & 0.83 & 1.02 & 0.89 & 1.03 & 1.00 & 0.81 & 0.80 & 0.96 & 1.01 & 0.40 & 0.22 & 0.87 & 0.36 & 1.04 \\
\midrule
\multirow{2}{*}{Gemini-2.5-Pro} & IO & 0.50 & 0.35 & 0.78 & 0.46 & 1.03 & 0.50 & 0.26 & 0.70 & 0.43 & 1.03 & 0.00 & -- & -- & -- & -- \\
& CoT & 0.60 & 0.37 & 0.90 & 0.50 & 1.01 & 0.50 & 0.26 & 0.76 & 0.45 & 0.98 & 0.50 & 0.27 & 0.84 & 0.41 & 1.03 \\
\midrule
\multirow{2}{*}{DeepSeek-V3.2-Exp} & IO & 1.00 & 0.85 & 0.84 & 0.94 & 1.00 & 0.70 & 0.36 & 0.86 & 0.56 & 1.03 & 0.20 & 0.18 & 0.66 & 0.30 & 1.02 \\
& CoT & 1.00 & 0.89 & 0.87 & 0.95 & 1.00 & 0.90 & 0.53 & 0.75 & 0.74 & 1.03 & 0.50 & 0.25 & 0.81 & 0.39 & 1.10 \\
\bottomrule
\end{tabular}
}
\caption{Normalized results relative to human performance within each difficulty level. All metrics are normalized such that human performance equals 1.0 for each difficulty separately, with higher values indicating better performance. For pOCT, nOCT, and pMD (originally lower-is-better metrics), inverse ratios are used.}
\label{tab:human_subset_results_normalized_per_difficulty}
\end{table*}

\section{Prompts}

\noindent
\textbf{Note on time constants.}
All time-related constants referenced in the prompts (e.g., \texttt{INTERACT\_TIME}, \texttt{PROCESS\_CUT\_TIME}) follow the fixed definitions summarized in Appendix \ref{sec:consts}.
These constants are explicitly provided to the LLM during planning.

\begin{prompt}[title={Prompt template for I/O Testing - Part 1}]
You are given the input map JSON, recipes, and orders, along with the Overcooked multi-agent parallel planning rules described below. Your goal is to generate a detailed action plan (Action List) for guiding each agent to complete dish preparation. The action plan must strictly follow the specified format and constraints.

Core Principles:
\ \ \ \ Maximize Efficiency: Minimize the total time required to complete all orders. This is the most critical goal.
\ \ \ \ Maximize Parallelism: Ensure multiple agents are working simultaneously whenever possible to reduce idle time.
\ \ \ \ Ensure Accuracy: Adhere 100\% to all action definitions, rules, and constraints outlined below.

Input Content:
\ \ \ \ Map JSON: Describes kitchen layout, station coordinates, initial items, and agent positions.
\ \ \ \ Recipes: Describes the preparation workflow and required ingredients for the dishes.
\ \ \ \ Orders: Describe the dishes that need to be completed in order.

Output Requirements:
\ \ \ \ For each agent, output an ordered action list (e.g., agent1, agent2).
\ \ \ \ Each action is a dictionary containing action type and parameters.
\ \ \ \ Please strictly follow the output standard JSON format action list. Do not add any additional explanations or content!

Output Format Example:
\ \ \ \ \{\{
\ \ \ \ \ \ \ \ "plan": \{\{
\ \ \ \ \ \ \ \ \ \ \ \ "agent1": [
\ \ \ \ \ \ \ \ \ \ \ \ \ \ \ \ \{\{"action": "MoveTo", "target": [x1, y1]\}\},
\ \ \ \ \ \ \ \ \ \ \ \ \ \ \ \ \{\{"action": "Interact", "target": "station\_name1"\}\},
\ \ \ \ \ \ \ \ \ \ \ \ \ \ \ \ ...
\ \ \ \ \ \ \ \ \ \ \ \ ],
\ \ \ \ \ \ \ \ \ \ \ \ "agent2": [
\ \ \ \ \ \ \ \ \ \ \ \ \ \ \ \ \{\{"action": "MoveTo", "target": [x2, y2]\}\},
\ \ \ \ \ \ \ \ \ \ \ \ \ \ \ \ \{\{"action": "Process", "target": "station\_name2"\}\},
\ \ \ \ \ \ \ \ \ \ \ \ \ \ \ \ ...
\ \ \ \ \ \ \ \ \ \ \ \ ],
\ \ \ \ \ \ \ \ \ \ \ \ ...
\ \ \ \ \ \ \ \ \ \ \ \}\}
\ \ \ \ \}\}

Task:

\{task\}

\end{prompt}

\begin{prompt}[title={Prompt template for I/O Testing - Part 2}]
Environment Rules and Constraints:

Agent Rules:
\ \ \ \ No Collision: Agents do not consider collision boxes between each other; their movement paths and positions can overlap at any time.
\ \ \ \ Single Item Hold: An agent can only hold one item at a time (e.g., an ingredient, a plate, a pot). Item exchange must be done via surfaces like tables; direct passing is not allowed. Cannot hold multiple ingredients or containers at once.
\ \ \ \ Positioning: Agents can only stand on empty floor tiles; actions must be performed on adjacent empty ground to target stations; movement can only occur through empty ground. At any time, an agent's coordinates can never overlap with a station's coordinates.
\ \ \ \ Agents can only interact or process with workstations that are adjacent in the four cardinal directions (up, down, left, right).

Environment \& Item Rules:
\ \ \ \ Station Exclusivity: Fixed stations like cutting boards or sinks can only be used by one agent at a time for a Process action.
\ \ \ \ Ingredient Dispensing: Ingredients can only be obtained from designated dispensers. Each dispenser provides a specific type of ingredient. All types of ingredients can be directly held without the need for additional containers.
\ \ \ \ Cooking Process:
\ \ \ \ \ \ \ \ Stoves can only hold cookware (pots/pans), not ingredients directly.
\ \ \ \ \ \ \ \ Cooking starts automatically once cookware is placed on a stove and contains ingredients. Picking it up pauses cooking; placing it back on any stove resumes it.
\ \ \ \ \ \ \ \ Cooked food cannot be picked up by hand; it must be transferred in a container.
\ \ \ \ Serving Process:
\ \ \ \ \ \ \ \ All food items must be placed on a plate before being submitted at the serving window. The order in which the ingredients are placed on the plate is not important.
\ \ \ \ \ \ \ \ Dishes must be served in the exact order specified in the Orders list.
\ \ \ \ Plate Cycle:
\ \ \ \ \ \ \ \ Dirty plates return to the dirty plate return station some time after a dish is served.
\ \ \ \ \ \ \ \ A dirty plate cannot hold any items and must be washed at a sink to become a clean plate.
\ \ \ \ Time Consumption:
\ \ \ \ \ \ \ \ Move: 1 unit per tile
\ \ \ \ \ \ \ \ Interact: {INTERACT\_TIME} units
\ \ \ \ \ \ \ \ Chopping: {PROCESS\_CUT\_TIME} units
\ \ \ \ \ \ \ \ Pot Cooking: {PROCESS\_POT\_COOK\_TIME} units
\ \ \ \ \ \ \ \ Pan Cooking: {PROCESS\_PAN\_COOK\_TIME} units
\ \ \ \ \ \ \ \ Washing Plates: {PROCESS\_WASH\_PLATE\_TIME} units
\ \ \ \ \ \ \ \ Dirty Plate Return: {RETURN\_DIRTY\_PLATE\_TIME} units

\end{prompt}

\begin{prompt}[title={Prompt template for I/O Testing - Part 3}]

Action Definitions:
\ \ \ \ MoveTo(coordinate):
\ \ \ \ \ \ \ \ format: \{\{"action": "MoveTo", "target": [x, y]\}\}
\ \ \ \ Interact(target\_name):
\ \ \ \ \ \ \ \ format: \{\{"action": "Interact", "target": "station\_name"\}\}
\ \ \ \ Process(target\_name):
\ \ \ \ \ \ \ \ format: \{\{"action": "Process", "target": "station\_name"\}\}
\ \ \ \ Wait(duration):
\ \ \ \ \ \ \ \ format: \{\{"action": "Wait", "duration": t\}\}
\ \ \ \ Finish():
\ \ \ \ \ \ \ \ format: \{\{"action": "Finish"\}\}

Suggestions:
\ \ \ \ Tasks must be reasonably allocated to achieve multi-agent parallel collaboration and minimize total time consumption.
\ \ \ \ Action sequence must completely cover the entire process from raw material acquisition, processing, assembly to serving.
\ \ \ \ Always notice the timepoint when each action starts and ends to ensure no conflicts in agent actions and get the most efficient plan.

\end{prompt}

\begin{prompt}[title={Prompt template for CoT Testing - Part 1}]
You are given the input map JSON, recipes, and orders, along with the Overcooked multi-agent parallel planning rules described below. Your goal is to generate a **step-by-step reasoning process** (Chain-of-Thought, CoT) that leads to a detailed action plan for guiding each agent to complete dish preparation. The reasoning must explicitly explain the allocation of subtasks, parallel coordination, and timing decisions. 

Core Principles:
\ \ \ \ Maximize Efficiency: Minimize the total time required to complete all orders. This is the most critical goal.
\ \ \ \ Maximize Parallelism: Ensure multiple agents are working simultaneously whenever possible to reduce idle time.
\ \ \ \ Ensure Accuracy: Adhere 100\% to all action definitions, rules, and constraints outlined below.

Input Content:
\ \ \ \ Map JSON: Describes kitchen layout, station coordinates, initial items, and agent positions.
\ \ \ \ Recipes: Describes the preparation workflow and required ingredients for the dishes.
\ \ \ \ Orders: Describe the dishes that need to be completed in order.

Output Requirements:
\ \ \ \ For each agent, output an ordered action list alongside the CoT reasoning steps.
\ \ \ \ Each step should include: reasoning about which subtask to execute, dependencies, and timing considerations.
\ \ \ \ Each action is a dictionary containing action type and parameters.
\ \ \ \ Please strictly follow the output standard JSON format for action lists and reasoning steps. Do not add any additional explanations outside of the CoT reasoning.

Output Format Example:
\ \ \ \ \{\{
\ \ \ \ \ \ \ \ "CoT": [
\ \ \ \ \ \ \ \ \ \ \ \ "Step 1: Agent1 moves to ingredient dispenser to pick up tomato, reasoning: starting first ingredient to minimize idle time",
\ \ \ \ \ \ \ \ \ \ \ \ "Step 2: Agent2 moves to counter to prepare plate, reasoning: parallel work to maximize efficiency",
\ \ \ \ \ \ \ \ \ \ \ \ ...
\ \ \ \ \ \ \ \ ],
\ \ \ \ \ \ \ \ "plan": \{\{
\ \ \ \ \ \ \ \ \ \ \ \ "agent1": [
\ \ \ \ \ \ \ \ \ \ \ \ \ \ \ \ \{\{"action": "MoveTo", "target": [x1, y1]\}\},
\ \ \ \ \ \ \ \ \ \ \ \ \ \ \ \ \{\{"action": "Interact", "target": "station\_name1"\}\},
\ \ \ \ \ \ \ \ \ \ \ \ \ \ \ \ ...
\ \ \ \ \ \ \ \ \ \ \ \ ],
\ \ \ \ \ \ \ \ \ \ \ \ "agent2": [
\ \ \ \ \ \ \ \ \ \ \ \ \ \ \ \ \{\{"action": "MoveTo", "target": [x2, y2]\}\},
\ \ \ \ \ \ \ \ \ \ \ \ \ \ \ \ \{\{"action": "Process", "target": "station\_name2"\}\},
\ \ \ \ \ \ \ \ \ \ \ \ \ \ \ \ ...
\ \ \ \ \ \ \ \ \ \ \ \ ],
\ \ \ \ \ \ \ \ \ \ \ \ ...
\ \ \ \ \ \ \ \ \ \ \ \}\}
\ \ \ \ \}\}

Task:

\{task\}

\end{prompt}

\begin{prompt}[title={Prompt template for CoT Testing - Part 2}]
Environment Rules and Constraints:

Agent Rules:
\ \ \ \ No Collision: Agents do not consider collision boxes between each other; movement paths and positions can overlap.
\ \ \ \ Single Item Hold: Agents can only hold one item at a time. Exchanges must be done via surfaces; direct passing is not allowed.
\ \ \ \ Positioning: Agents can only stand on empty floor tiles; movement can only occur through empty ground; coordinates cannot overlap with stations.
\ \ \ \ Interactions: Only with adjacent workstations in four cardinal directions.

Environment \& Item Rules:
\ \ \ \ Station Exclusivity: Fixed stations can only be used by one agent at a time.
\ \ \ \ Ingredient Dispensing: Ingredients obtained only from designated dispensers.
\ \ \ \ Cooking Process: Stoves hold cookware, start automatically when ingredients are inside; cooked food must be transferred in a container.
\ \ \ \ Serving Process: Food must be plated before submission; served in order of Orders list.
\ \ \ \ Plate Cycle: Dirty plates return to the dirty plate return station; washed at a sink to become clean.
\ \ \ \ Time Costs:
\ \ \ \ \ \ \ \ Move: 1 unit per tile
\ \ \ \ \ \ \ \ Interact: {INTERACT\_TIME} units
\ \ \ \ \ \ \ \ Chopping: {PROCESS\_CUT\_TIME} units
\ \ \ \ \ \ \ \ Pot Cooking: {PROCESS\_POT\_COOK\_TIME} units
\ \ \ \ \ \ \ \ Pan Cooking: {PROCESS\_PAN\_COOK\_TIME} units
\ \ \ \ \ \ \ \ Washing Plates: {PROCESS\_WASH\_PLATE\_TIME} units
\ \ \ \ \ \ \ \ Dirty Plate Return: {RETURN\_DIRTY\_PLATE\_TIME} units

Action Definitions:
\ \ \ \ MoveTo(coordinate):
\ \ \ \ \ \ \ \ format: \{\{"action": "MoveTo", "target": [x, y]\}\}
\ \ \ \ Interact(target\_name):
\ \ \ \ \ \ \ \ format: \{\{"action": "Interact", "target": "station\_name"\}\}
\ \ \ \ Process(target\_name):
\ \ \ \ \ \ \ \ format: \{\{"action": "Process", "target": "station\_name"\}\}
\ \ \ \ Wait(duration):
\ \ \ \ \ \ \ \ format: \{\{"action": "Wait", "duration": t\}\}
\ \ \ \ Finish():
\ \ \ \ \ \ \ \ format: \{\{"action": "Finish"\}\}

Suggestions:
\ \ \ \ Tasks must be reasonably allocated to achieve multi-agent parallel collaboration and minimize total time consumption.
\ \ \ \ Action sequence must completely cover the entire process from raw material acquisition, processing, assembly to serving.
\ \ \ \ Always notice the timepoint when each action starts and ends to ensure no conflicts in agent actions and get the most efficient plan.

\end{prompt}

\section{Environment Constants}
\label{sec:consts}

In ParaCook, LLM agents are given the execution time of each action during planning.
All constants listed below are shared across all environments and models.

\begin{table}[h]
\centering
\small
\resizebox{0.48\textwidth}{!}{
\begin{tabular}{lll}
\toprule
Constant Name & Action / Process & Time Cost \\
\midrule
\texttt{MOVE\_TIME} & Move per tile & 1 \\
\texttt{INTERACT\_TIME} & Pickup / Drop & 0 \\
\texttt{PROCESS\_CUT\_TIME} & Cut & 4 \\
\texttt{PROCESS\_POT\_COOK\_TIME} & Cook (Pot) & 16 \\
\texttt{PROCESS\_PAN\_COOK\_TIME} & Cook (Pan) & 24 \\
\texttt{PROCESS\_WASH\_PLATE\_TIME} & Wash plate & 6 \\
\texttt{RETURN\_DIRTY\_PLATE\_TIME} & Return dirty plate & 10 \\
\bottomrule
\end{tabular}
}
\caption{Time constants in ParaCook.}
\end{table}

\section{Menu Recipes}

This section provides a comprehensive overview of all recipes in the dataset, organized by food category. The dataset contains 20 recipes across 6 categories: Burger, Burrito, Pasta, Salad, Sashimi, and Sushi. Each recipe includes detailed preparation instructions.

\subsection{Burger}

Burgers are a classic dish consisting of cooked meat patties served with bread. The dataset includes five burger variations, ranging from basic burgers to more elaborate versions with additional toppings such as lettuce, tomato, and cheese. All burger recipes require cooking the meat in a pan and assembling the ingredients on a plate.

\paragraph{Basic Burger (burger\_basic)}
First chop the meat and cook it in pan. Then put the cooked meat with a piece of bread on a plate to make a basic burger.

\paragraph{Burger with Lettuce (burger\_lettuce)}
Chop the meat and cook it in pan. Chop the lettuce. Then put the cooked meat, chopped lettuce with a piece of bread on a plate to make a burger with lettuce.

\paragraph{Full Burger (burger\_full)}
Chop the meat and cook it in pan. Chop the lettuce and tomato. Then put the cooked meat, chopped lettuce, chopped tomato with a piece of bread on a plate to make a full burger.

\paragraph{Burger with Cheese (burger\_cheese)}
Chop the meat and cook it in pan. Then put the cooked meat with a piece of bread and a slice of cheese on a plate to make a burger with cheese.

\paragraph{Burger with Cheese and Lettuce (burger\_cheese\_lettuce)}
Chop the meat and cook it in pan. Chop the lettuce. Then put the cooked meat, chopped lettuce with a piece of bread and a slice of cheese on a plate to make a burger with cheese and lettuce.

\subsection{Burrito}

Burritos are Mexican-inspired dishes that combine cooked rice with protein, wrapped in a tortilla. The dataset features three burrito variations with different protein options: meat, chicken, and mushroom. All burrito recipes require cooking rice in a pot and the protein in a pan before assembly.

\paragraph{Burrito with Meat (burrito\_meat)}
Chop and cook the meat in pan, cook the rice in pot, then put cooked meat and cooked rice together with a raw tortilla to a plate to make a burrito with meat.

\paragraph{Burrito with Chicken (burrito\_chicken)}
Chop and cook the chicken in pan, cook the rice in pot, then put cooked chicken and cooked rice together with a raw tortilla to a plate to make a burrito with chicken.

\paragraph{Burrito with Mushroom (burrito\_mushroom)}
Chop and cook the mushroom in pan, cook the rice in pot, then put cooked mushroom and cooked rice together with a raw tortilla to a plate to make a burrito with mushroom.

\subsection{Pasta}

Pasta dishes combine cooked pasta with various sauces and toppings. The dataset includes four pasta variations featuring tomato, meat, mushroom, and seafood. All pasta recipes require cooking the pasta in a pot and preparing the sauce or protein in a pan.

\paragraph{Pasta with Tomato (pasta\_tomato)}
Cook the pasta in pot, chop the tomato and cook it in pan, then put cooked pasta and cooked tomato together to a plate to make pasta with tomato pasta.

\paragraph{Pasta with Meat (pasta\_meat)}
Cook the pasta in pot, chop the meat and cook it in pan, then put cooked pasta and cooked meat together to a plate to make pasta with meat sauce.

\paragraph{Pasta with Mushroom (pasta\_mushroom)}
Cook the pasta in pot, chop the mushroom and cook it in pan, then put cooked pasta and cooked mushroom together to a plate to make pasta with mushroom sauce.

\paragraph{Seafood Pasta (pasta\_seafood)}
Cook the pasta in pot, chop the fish and prawn and cook them in pan respectively, then put cooked pasta, cooked fish and cooked prawn together to a plate to make seafood pasta.

\subsection{Salad}

Salads are fresh vegetable dishes that require no cooking. The dataset contains three salad variations of increasing complexity, from a basic lettuce salad to a full salad with multiple vegetables. All salad recipes only require chopping vegetables and plating them.

\paragraph{Basic Salad (salad\_basic)}
Put chopped lettuce on a plate to make a salad.

\paragraph{Advanced Salad (salad\_advanced)}
Put chopped lettuce and chopped tomato together to a plate to make a salad.

\paragraph{Full Salad (salad\_full)}
Put chopped lettuce, chopped tomato and chopped cucumber together to a plate to make a salad.

\subsection{Sashimi}

Sashimi is a Japanese dish featuring raw seafood that is simply chopped and plated. The dataset includes two sashimi variations: fish and shrimp. These are the simplest recipes in the dataset, requiring only chopping and plating.

\paragraph{Sashimi with Fish (sashimi\_fish)}
Chop the fish and put the chopped fish to a plate to make sashimi with fish.

\paragraph{Sashimi with Shrimp (sashimi\_shrimp)}
Chop the shrimp and put the chopped shrimp to a plate to make sashimi with shrimp.

\subsection{Sushi}

Sushi is a Japanese dish that combines cooked rice, nori (seaweed), and various fillings. The dataset features three sushi variations with fish, cucumber, or both. All sushi recipes require cooking rice in a pot and assembling with raw nori and chopped ingredients.

\paragraph{Fish Sushi (sushi\_fish)}
First chop the fish and cook the rice. Then put the chopped fish and cooked rice and a piece of nori on a plate to make a fish sushi.

\paragraph{Cucumber Sushi (sushi\_cucumber)}
First chop the cucumber and cook the rice. Then put the chopped cucumber and cooked rice and a piece of nori on a plate to make a cucumber sushi.

\paragraph{Full Sushi (sushi\_full)}
First chop the fish and cucumber and cook the rice. Then put the chopped fish, chopped cucumber and cooked rice and a piece of nori on a plate to make a full sushi.

\section{Dataset Extensibility}
ParaCook is designed not as a fixed dataset but as a flexible framework. Both tasks and map configuration support extensions. New recipes and cooking tools can be added to create richer workflows, while recipe complexity, order size, and resource constraints can be systematically varied. Maps can be scaled or altered during execution, and stochastic events (e.g., overcooking, fires) can be introduced to test robustness and adaptive planning. 
In addition, Maps also supports particular collaborative planning with specific designs.
For instance, a row of tables may divide the kitchen, placing different types of workstations on each side, so that agents on each side can only perform certain tasks and must pass items through the central tables for coordination.
This extensibility keeps the framework useful for evaluating complex planning as agents advance.

\section{Derivation of Upper Bounds for Time and Distance}
\label{sec:upperbound}

The upper bounds $D_{\max}$ and $T_{\max}$ for an order are computed under a sequential execution assumption.

\subsection{Maximum Single-Step Movement Distance}

We define the maximum possible movement distance for a single navigation step as
\begin{equation}
d = H + W,
\end{equation}
where $H$ and $W$ denote the height and width of the grid map, respectively.
This corresponds to the Manhattan distance between the top-left and bottom-right corners of the map.
Any single movement in the environment is upper-bounded by $d$.

\subsection{Upper Bound for a Single Dish}

For each dish, we compute an upper bound on both movement distance and completion time by traversing all required ingredients and their processing states.
For each ingredient, the following costs are accumulated:

\begin{itemize}
    \item \textbf{Raw}: movement distance $d$ to fetch the ingredient from the dispenser.
    \item \textbf{Chopped}: movement distance $d$ to reach the cutting board, plus cutting time.
    \item \textbf{Cooked}: movement distance $d$ to reach the cooking station, plus cooking time.
    \item Movement distance $d$ to place the processed ingredient onto a plate.
\end{itemize}

After all ingredients are processed, an additional movement distance $d$ is added to deliver the dish to the serving window.

\subsection{Upper Bound for an Order}

Given an order consisting of multiple dishes, we compute the overall upper bounds $D_{\max}$ and $T_{\max}$ by summing the corresponding bounds of all dishes, assuming sequential execution of all actions.

In addition, we account for plate reuse.
Each map initially provides $m$ clean plates.
If the number of dishes in an order exceeds $m$, each additional dish incurs the following extra costs:

\begin{itemize}
    \item Movement distance $d$ to reach the plate return location.
    \item Waiting time for dirty plate return: \texttt{RETURN\_DIRTY\_PLATE\_TIME}.
    \item Movement distance $d$ to reach the sink.
    \item Plate washing time: \texttt{PROCESS\_WASH\_PLATE\_TIME}.
\end{itemize}

\section{Evaluation Configurations}
\label{sec:eval_config}

\begin{table}[h]
\centering
\small
\resizebox{0.45\textwidth}{!}{
\begin{tabular}{ll}
\toprule
Factor & Values \\
\midrule
Recipes & 6 \\
Order sizes & 1, 2, 3, 4 \\
Number of agents & 1, 2, 3 \\
Map seeds & 42, 84, 126, 128, 256 \\
\midrule
Total instances & $6 \times 4 \times 3 \times 5 = 360$ \\
\bottomrule
\end{tabular}
}
\caption{Enumeration of evaluation configurations used in all experiments.}
\label{tab:eval_config}
\end{table}

Each configuration corresponds to a unique task instance instantiated by a fixed recipe, order size, number of agents, and map seed.

\section{Time-Budget Decomposition of Execution}
\label{sec:time_breakdown}

To understand factors affecting completion time, we decompose successful run execution time into \textbf{movement}, \textbf{processing}, and \textbf{waiting} components. Table \ref{tab:time_breakdown} presents GPT-5 (CoT) results, which are the best-performing model, along with human performance as a reference. 

Across all difficulty levels, movement time constitutes the largest portion of total completion time. The processing time remains relatively small and varies primarily with the recipe composition. Waiting time is consistently lower than movement time and comparable to or slightly larger than processing time, reflecting delays induced by task dependencies and resource availability.

A similar distribution pattern is observed for human performance, where movement also dominates the time budget, but with consistently lower values. This suggests that both humans and LLMs operate under the same execution constraints, while differing in their efficiency in spatial planning and coordination.

\begin{table}[t]
\centering
\small
\resizebox{0.48\textwidth}{!}{
\begin{tabular}{llcccc}
\toprule
\textbf{Model} & \textbf{Difficulty} & \textbf{Move} & \textbf{Wait} & \textbf{Process} & \textbf{OCT} \\
\midrule
GPT-5 (CoT) & Easy   & 43.15 & 2.14  & 9.27 & 54.55 \\
GPT-5 (CoT) & Medium & 75.38 & 10.33 & 7.00 & 92.71 \\
GPT-5 (CoT) & Hard   & 74.25 & 4.25  & 9.63 & 88.13 \\
\midrule
Human & Easy   & 37.69 & 4.25 & 7.25 & 49.19 \\
Human & Medium & 64.50 & 7.44 & 5.50 & 77.44 \\
Human & Hard   & 60.56 & 5.75 & 7.25 & 73.56 \\
\bottomrule
\end{tabular}
}
\caption{Time-budget decomposition of successful runs across difficulty levels.}
\label{tab:time_breakdown}
\end{table}

\section{Error Analysis}
\label{sec:error_analysis}

To better understand the concrete planning failures exposed by ParaCook, we analyze execution error logs generated during evaluation.
We find that failures predominantly arise from semantic hallucinations and violations of environmental preconditions during the execution of plans. Specifically, we identified three major error patterns in the generated plans.

\paragraph{Infeasible locations}
A common failure mode is that agents are instructed to move to infeasible locations.
In particular, some generated plans direct agents to grid cells that are already occupied by workstations.
According to the environment specification, such positions are invalid and cannot be entered.
These errors indicate that, during planning, models may fail to consistently respect strict spatial constraints encoded in the map, even when such constraints are explicitly provided in the prompt.

\paragraph{Non-adjacent interaction}
Another frequent error arises from interacting with workstations from non-adjacent positions.
The environment requires agents to be located in one of the four adjacent cells to perform interaction actions.
However, some plans attempt to execute interactions while the agent is not adjacent to the target workstation, violating action preconditions and leading to execution failures.
This suggests that models may lose track of fine-grained spatial relations between agents and objects over the course of multi-step planning.

\paragraph{Holding-item violations}
We also observe failures caused by violations of object-holding constraints.
Certain actions, such as picking up ingredients or tools, require the agent to have free hands.
Nevertheless, some generated plans issue pickup actions when the agent is already holding an item.
As a result, the execution enters inconsistent states, and subsequent actions are taken under incorrect assumptions about the agent's inventory, causing cascading failures later in the plan.

These error patterns help explain the performance trends observed in the experimental results.
On simpler tasks with shorter plans, such semantic violations may not accumulate, allowing occasional successful executions.
In contrast, as task complexity and planning horizon increase, maintaining consistent state awareness and respecting environmental constraints becomes increasingly difficult, leading to sharp performance degradation on medium and hard tasks.

\section{Advanced Planning Methods}
\label{sec:advanced_methods}

To address whether structured planning architectures can improve performance on ParaCook, we evaluate two additional methods beyond vanilla IO and CoT prompting: PLaG (graph-augmented prompting) and MultiStepReAct (interactive planning with environmental feedback). Due to computational cost considerations, we evaluate PLaG and MultiStepReAct on a representative subset of ParaCook tasks.

\subsection{Method Descriptions}

\paragraph{PLaG (Plan-like-a-Graph)} PLaG \citep{lin2024graphenhancedlargelanguagemodels} explicitly constructs a task dependency graph before action generation. The LLM first decomposes tasks into subtasks with dependency relations and estimates the duration of each subtask, then generates action sequences that respect these constraints. This makes temporal dependencies, task durations, and parallelization opportunities explicit.

\paragraph{MultiStepReAct (Multi-Step ReAct)} MultiStepReAct interleaves planning and execution iteratively. At each step, the model observes the current state, plans the next batch of actions, executes them, and receives feedback before continuing. This differs from standard ReAct, which plans one action at a time. Such an approach would incur prohibitively high costs given ParaCook's long action sequences. MultiStepReAct balances upfront planning with reactive adaptation.

\subsection{Results and Analysis}

Table~\ref{tab:advanced_methods_results} presents the results across all models and difficulty levels.

\textbf{PLaG shows model-dependent effects with no consistent improvements.} The impact of explicit graph construction varies significantly across models. For GPT-5, PLaG achieves 72.92\% on Easy, 62.50\% on Medium, and 54.17\% on Hard, comparable to or slightly below CoT performance (84.17\%, 77.03\%, 57.39\%). DeepSeek shows similar patterns with 58.33\%, 60.42\%, and 35.42\% across difficulties. In contrast, Gemini degrades notably with PLaG, dropping to 18.75\% on Hard tasks. Claude shows modest improvements over its IO/CoT baseline, rising from 0\% to 8.33\% on Hard tasks. These mixed results suggest that for models with strong implicit reasoning capabilities, forcing an explicit graph structure may introduce additional complexity that interferes with their native planning process. Weaker models like Claude may benefit from the structural scaffolding, though their absolute performance remains limited.

\textbf{MultiStepReAct substantially benefits weaker models but shows mixed or negative effects on others.} Claude shows dramatic improvements with interactive planning, rising from 55.00\% (CoT) to 95.83\% (MultiStepReAct) on Easy tasks, even surpassing GPT-5's 84.17\%. However, this advantage diminishes on harder tasks, where Claude achieves only 25\% compared to GPT-5's 65.22\%. For other models, MultiStepReAct shows inconsistent or negative effects: GPT-5 maintains similar performance (83.33\% Easy, 59.09\% Medium, 65.22\% Hard), while Gemini degrades on Medium (37.50\% vs. CoT 47.22\%) and Hard (20.83\% vs. 37.50\%), and DeepSeek drops substantially on Easy (52.38\% vs. CoT 67.50\%). This suggests that environmental feedback can compensate for weaker intrinsic planning capabilities through iterative refinement, but may introduce overhead or distraction for models that already possess strong reasoning abilities.

\textbf{Time efficiency shows limited improvements and method-dependent patterns.} PLaG generally does not improve execution efficiency, with nOCT remaining comparable to or higher than baseline across most cases. For example, GPT-5 achieves nOCT of 27.56, 24.06, and 22.77 across difficulties, similar to its CoT performance. MultiStepReAct exhibits more varied patterns: Claude benefits substantially on Easy (32.42 vs. CoT 33.20) and Medium (27.73 vs. 37.61) tasks, indicating that feedback helps translate plans into efficient execution. However, other models show mixed results. GPT-5's nOCT slightly increases on Easy (30.39 vs. CoT 26.84), and Gemini shows minimal change or degradation. Overall, neither method consistently improves time efficiency across models and difficulties.

\textbf{The core bottleneck is grounding high-level plans into executable actions.} PLaG aids high-level decomposition but does not address how to execute actions under environmental constraints such as spatial positioning, resource contention, and timing dependencies. MultiStepReAct provides execution feedback that helps weaker models adapt, but becomes insufficient as complexity increases. Neither approach provides a universal solution, confirming that ParaCook exposes challenges beyond pure scheduling optimization and suggesting that future work should explore hybrid approaches combining parallel scheduling with constraint-aware grounding mechanisms.

\begin{table*}[t]
\centering
\small
\resizebox{\textwidth}{!}{
\begin{tabular}{l l ccccc ccccc ccccc}
\toprule
& & \multicolumn{5}{c}{Easy} & \multicolumn{5}{c}{Medium} & \multicolumn{5}{c}{Hard} \\
\cmidrule(lr){3-7} \cmidrule(lr){8-12} \cmidrule(lr){13-17}
Model & Method & SR$\uparrow$ & pOCT$\downarrow$ & nOCT$\downarrow$ & pMD$\downarrow$ & AU$\uparrow$ & SR$\uparrow$ & pOCT$\downarrow$ & nOCT$\downarrow$ & pMD$\downarrow$ & AU$\uparrow$ & SR$\uparrow$ & pOCT$\downarrow$ & nOCT$\downarrow$ & pMD$\downarrow$ & AU$\uparrow$ \\
\midrule
\multirow{2}{*}{GPT-5} & PLaG & 72.92 & 143.25 & 27.56 & 69.52 & 91.43 & 62.50 & 354.58 & 24.06 & 152.86 & 86.35 & 54.17 & 370.69 & 22.77 & 165.33 & 90.60 \\
                       & MSReAct & 83.33 & 139.71 & 30.39 & 72.81 & 86.62 & 59.09 & 393.59 & 24.88 & 176.49 & 75.12 & 65.22 & 360.35 & 24.00 & 161.00 & 84.27 \\
\midrule
\multirow{2}{*}{Gemini-2.5-Pro} & PLaG & 54.17 & 193.46 & 31.54 & 90.02 & 93.12 & 47.92 & 412.33 & 24.45 & 176.98 & 82.71 & 18.75 & 502.71 & 35.79 & 220.39 & 99.35 \\
                                & MSReAct & 79.17 & 162.71 & 32.06 & 79.81 & 91.18 & 37.50 & 484.83 & 28.90 & 200.08 & 81.44 & 20.83 & 526.46 & 28.19 & 232.31 & 84.26 \\
\midrule
\multirow{2}{*}{DeepSeek-V3.2-Exp} & PLaG & 58.33 & 192.60 & 29.79 & 89.06 & 89.03 & 60.42 & 356.50 & 25.60 & 152.78 & 87.70 & 35.42 & 466.73 & 27.58 & 204.82 & 91.48 \\
                                   & MSReAct & 52.38 & 210.33 & 35.81 & 96.62 & 91.12 & 40.91 & 398.05 & 28.28 & 176.43 & 94.87 & 34.78 & 460.17 & 26.41 & 202.28 & 87.44 \\
\midrule
\multirow{2}{*}{Claude-Opus-4.1} & PLaG & 50.00 & 216.85 & 36.24 & 99.01 & 81.13 & 16.67 & 545.94 & 33.90 & 220.69 & 68.39 & 8.33 & 541.81 & 28.89 & 235.88 & 79.62 \\
                                 & MSReAct & 95.83 & 109.42 & 32.42 & 66.00 & 91.46 & 50.00 & 131.80 & 27.73 & 73.94 & 67.89 & 25.00 & 442.11 & 27.23 & 194.78 & 81.99 \\
\midrule
\multirow{2}{*}{Qwen3-Max-Preview} & PLaG & 29.17 & 255.43 & 26.04 & 116.43 & 76.24 & 6.94 & 587.22 & 28.21 & 237.51 & 84.69 & 2.78 & 570.01 & 32.17 & 248.62 & 75.78 \\
                                   & MSReAct & 41.67 & 231.79 & 21.79 & 106.33 & 75.70 & 8.33 & 622.29 & 33.16 & 251.06 & 100.00 & 4.35 & 610.22 & 22.87 & 267.20 & 100.00 \\
\bottomrule
\end{tabular}
}
\caption{Results of PLaG and MultiStepReAct (MSReAct) on state-of-the-art models across three difficulty levels. PLaG refers to Plan-like-a-Graph with explicit dependency graph construction. MSReAct refers to Multi-Step Reactive Planning with iterative planning-execution cycles.}
\label{tab:advanced_methods_results}
\end{table*}

\section{Discussion on Classical OR Solvers in Embodied Tasks}
\label{sec:or_discussion}

Classical OR solvers are highly effective for high-level scheduling where task parameters are deterministic and discrete. This is why we successfully incorporated a CP-SAT optimal solver into our abstract suite to establish performance upper bounds. However, applying these methods to the Embodied ParaCook tasks presents a fundamental modeling gap.

The primary difficulty lies in the fact that ParaCook’s embodied environment cannot be easily reduced to a static symbolic form for classical solvers due to several factors:
\begin{itemize}
    \item \textbf{Dynamic Spatial-Temporal Coupling:} Unlike standard scheduling, where the time spent on each step is often fixed, travel times in ParaCook depend on real-time navigation in a grid map. These are dynamically influenced by the instantaneous positions of agents and workstations, making it difficult to pre-calculate a deterministic distance matrix.
    \item \textbf{Difficulty of Abstraction:} 
    It required an exponential number of constraints to account for every possible spatial-temporal state with the MILP or STN framework, for example, modeling workstation exclusivity, inter-agent item flows, and interruptible operations.
    This leads to a severe ``state explosion'' when attempting to represent the full granularity of the embodied environment.
    \item \textbf{Generalizable Reasoning vs. Specialized Optimization:} Our benchmark is designed to evaluate LLMs' \textit{generalizable agentic reasoning}---specifically their ability to handle planning with grounding and execution in an environment that is hard to abstract. 
\end{itemize}

In summary, while OR solvers provide a theoretical limit for high-level logic, they are not directly applicable as baselines for the full embodied task, which requires the kind of dynamic, context-aware coordination that our benchmark aims to test in LLMs.

\section{Human Evaluation Interface}
\label{sec:human_gui}

\begin{figure*}[t]
\centering
\setlength{\belowcaptionskip}{-0.5cm}
\includegraphics[width=1\linewidth]{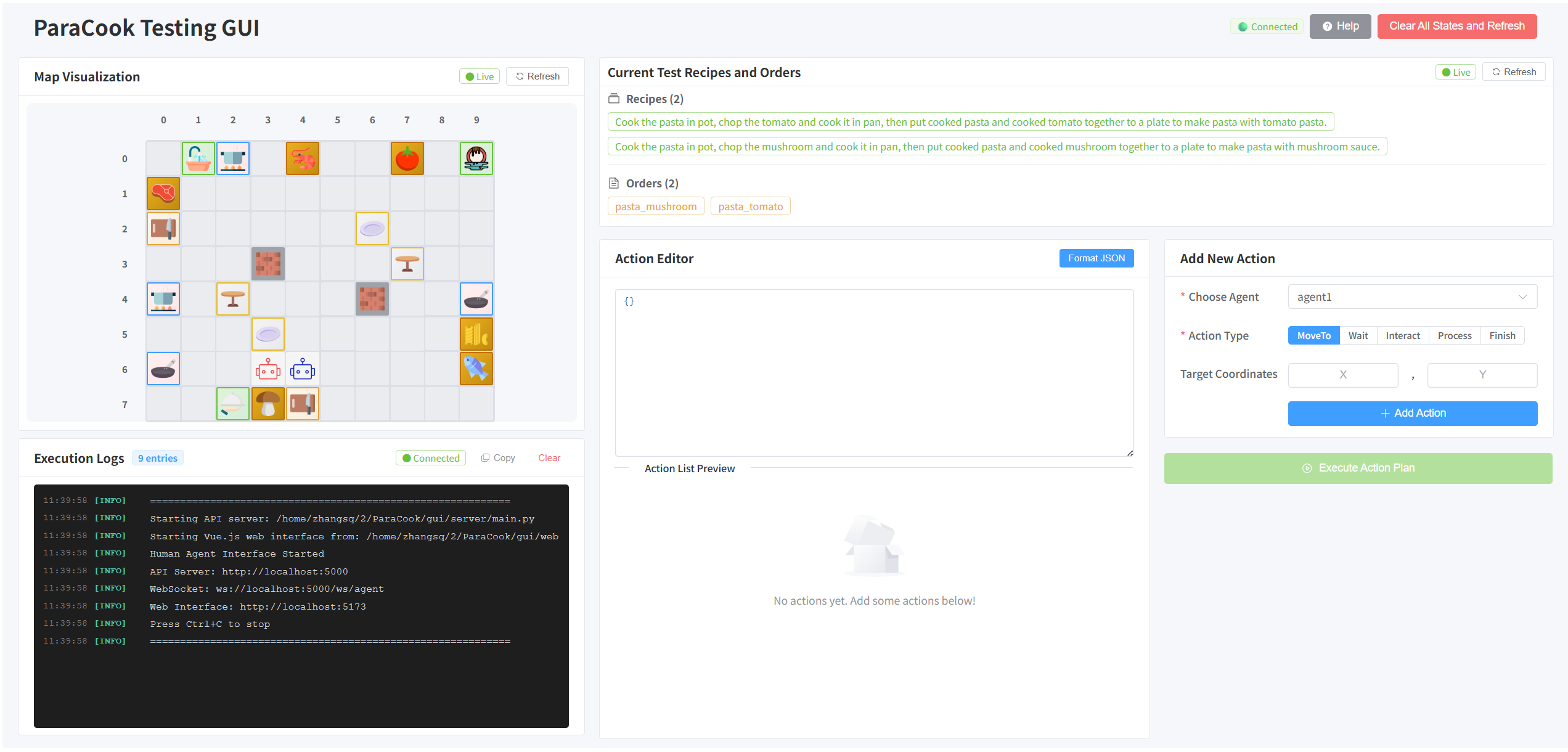}
\caption{The graphical user interface (GUI) used for human evaluation in ParaCook.
    The interface visualizes the environment state and allows participants to assign and execute action sequences for agents.}
\label{fig:human_gui}
\end{figure*}

To facilitate human evaluation, we design a graphical user interface (GUI) for interacting with the ParaCook environment.
Figure \ref{fig:human_gui} illustrates an example of the interface.

Human participants were provided with the same task description as the LLMs, including the map layout, available agents, recipes, and orders. No additional hints or strategy guidance were given. 
After reading the task instructions, human participants complete the task by assigning and executing action sequences for agents according to the given orders.
The GUI visualizes the map layout, agent positions, workstation states, and current orders to support human planning and coordination.

The GUI serves purely as an interaction layer for human usability.
All actions are executed under the same environment rules, constraints, and time model as those used for LLM-based evaluations, and the interface does not affect task dynamics or evaluation metrics.

\end{document}